\documentclass[10pt,journal,compsoc,twoside]{IEEEtran}

\usepackage{hyperref,cite}
\usepackage{url}
\usepackage{amsmath, amssymb} % Often used alongside amsthm for math symbols
\usepackage{amsthm}  % This is the crucial package
\usepackage{graphicx}
\usepackage{subcaption}
\usepackage{algorithm}
\usepackage{algpseudocode}
\usepackage[dvipsnames]{xcolor} 
\usepackage{booktabs} 
\usepackage{authblk}
\usepackage{float}
\usepackage{tikz}
\usepackage{colortbl}
\usetikzlibrary{arrows.meta, positioning, bending}

\newcommand*\circled[1]{\tikz[baseline=(char.base)]{
            \node[shape=circle,draw,inner sep=0.8pt] (char) {#1};}}

\definecolor{teal}{RGB}{0, 128, 128}
\definecolor{purple}{RGB}{128, 0, 128}
\definecolor{darkbg}{RGB}{20, 20, 40}

\newtheorem{proposition}{Proposition}[section]
\newtheorem{theorem}{Theorem}[section]

\begin{document}

% -------------------------------
%   TITLE INFORMATION
% -------------------------------
%\title{\textbf{Continual Learning through Structural Plasticity}}
\title{Task Switching Without Forgetting \\ via Proximal Decoupling}

\author{
    Pourya~Shamsolmoali,~\IEEEmembership{Senior Member,~IEEE,} 
    Masoumeh Zareapoor,~\IEEEmembership{Member,~IEEE,}
    Eric Granger,~\IEEEmembership{Member,~IEEE,}
    William A. P. Smith
    and Yue Lu,~\IEEEmembership{Senior Member,~IEEE}% <-this % stops a space

    \thanks{P.~Shamsolmoali and W.~Smith are with the Department of Computer Science, University of York, UK (\{pshams55@gmail.com, william.smith@york.ac.uk)}
\thanks{M.~Zareapoor is with the SEIEE, Shanghai Jiao Tong University, China (mzarea222@gmail.com).}
\thanks{E.~Granger is with LIVIA, Dept. of Systems Engineering, ETS Montreal, Canada (eric.granger@etsmtl.ca).}
\thanks{Y.~Lu is with the School of Communication and Electronic Eng., East China Normal University, China (ylu@cee.ecnu.edu.cn).}

}
% \IEEEcompsocitemizethanks{\IEEEcompsocthanksitem P.~Shamsolmoali and W. A. P. Smith are with the Department of Computer Science, University of York, York, United Kingdom. \protect\\
% E-mail: \{pourya.shamsolmoali, william.smith\}@york.ac.uk
% \IEEEcompsocthanksitem M.~Zareapoor, E.~Granger and Y.~Lu are with ...
% }
% }

% \affil{
%     {University of York,} \; {ETS Montreal,}\; {East China Normal University}
% }
% -------------------------------

% The paper headers
\markboth{IEEE TRANSACTIONS ON PATTERN ANALYSIS AND MACHINE INTELLIGENCE,~Vol.~X, No.~X, December~2025}%
{Shamsolmoali \MakeLowercase{\textit{et al.}} Task Switching Without Forgetting: The Architecture of Sequential Learning}

\IEEEtitleabstractindextext{%
\begin{abstract}
In continual learning, the primary challenge is to learn new information without forgetting old knowledge. A common solution addresses this trade-off through regularization, penalizing changes to parameters critical for previous tasks. In most cases, this regularization term is directly added to the training loss and optimized with standard gradient descent, which blends learning and retention signals into a single update and does not explicitly separate essential parameters from redundant ones. As task sequences grow, this coupling can over-constrain the model, limiting forward transfer and leading to inefficient use of capacity. We propose a different approach that separates task learning from stability enforcement via operator splitting. The learning step focuses on minimizing the current task loss, while a proximal stability step applies a sparse regularizer to prune unnecessary parameters and preserve task-relevant ones. This turns the stability-plasticity into a negotiated update between two complementary operators, rather than a conflicting gradient. We provide theoretical justification for the splitting method on the continual-learning objective, and demonstrate that our proposed solver achieves state-of-the-art results on standard benchmarks, improving both stability and adaptability without the need for replay buffers, Bayesian sampling, or meta-learning components.
\end{abstract}

}

\maketitle

\IEEEdisplaynontitleabstractindextext
\IEEEpeerreviewmaketitle

%%%%%%%%%%%%%%%%%%%%%%%
%%%%%%%%%%%%%%%%%%%%%%%
\IEEEraisesectionheading{\section{Introduction}}
\label{sec:intro}

\begin{figure*}
    \centering
    \includegraphics[width=0.86\textwidth]{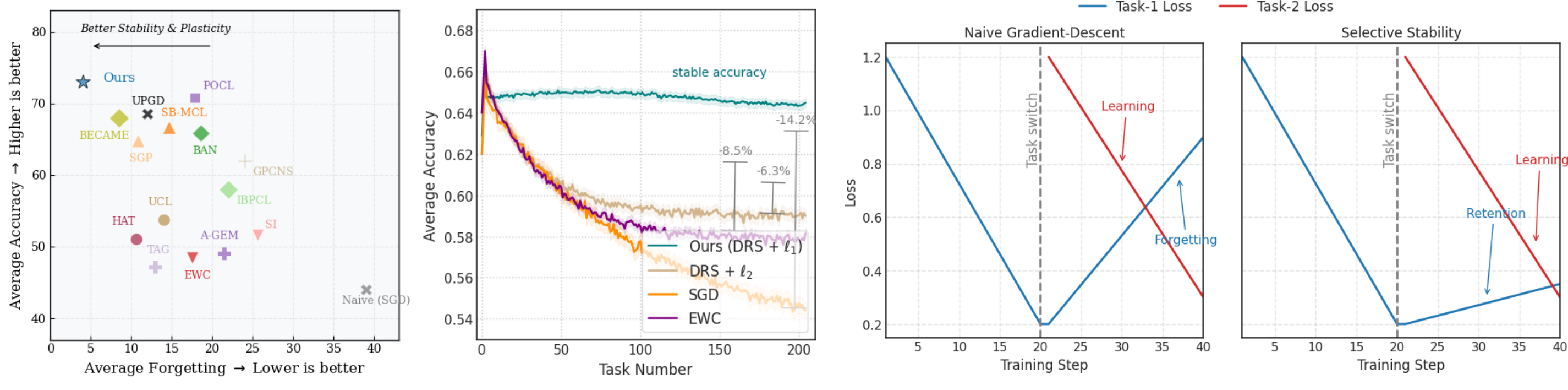} 
    \caption{Stability-plasticity trade-off in continual learning. Left: Average accuracy vs. average forgetting on the label-permuted EMNIST. An ideal model achieves both higher accuracy and lower forgetting. Middle: Long-term stability on input-permuted CIFAR-10, where SGD and EWC gradually degrade as forgetting accumulates (dropping $>8-13\%$). Right:  Illustration of the plasticity-stability dilemma. Under naive gradient descent, learning a new task (red) increases the loss on the old task (blue). With selective stability, the old task's loss remains low while the new task is learned. } \label{fig1}
\end{figure*}

\IEEEPARstart{I}n many real-world settings, learning is sequential rather than static. For example, an image recognition system may need to add new object categories over time without revisiting earlier data.  Continual learning (CL) formalizes this setting by training models on a stream of tasks while aiming to preserve performance on previous ones. A central challenge in this setting is catastrophic forgetting, where updates for new tasks degrade earlier knowledge \cite{mccloskey1989catastrophic, aljundi2018memory, dong2023no, luo2025theoretical, french1999catastrophic}. This behavior occurs because parameter updates that support new learning can interfere with parameters that are important for previous tasks, producing the well-known stability-plasticity dilemma (Fig. \ref{fig1}(c)). In essence, a model must remain plastic enough to learn new knowledge, while stable enough to preserve what it has already learned. To address this challenge, a broad set of CL strategies have been proposed, each introducing trade-offs that limit robustness and scalability. 
Memory-based (rehearsal) methods \cite{wu2024mitigating,yoo2024layerwise, elsayed2024addressing, thapabayesian,eskandar2025star, chaudhry2018efficient} mitigate forgetting by storing or synthesizing samples from previous tasks and replaying them during training. While effective at preserving past knowledge, their memory and computational cost grows with the number of tasks.
Architectural methods \cite{rusu2016progressive, saha2023continual, lyle2024disentangling} add task-specific modules or subnetworks, which prevents interference with past tasks, but leads to growth in model size, making it impractical for long-term continual learning. 
Meta-learning or hybrid approaches \cite{javed2019meta, kemker2018measuring, lee2024recasting, lee2024learning} frame CL as a bi-level optimization problem, learning update rules that generalize across tasks. These methods can improve adaptability, but they rely on expensive meta-training process \cite{li2025became}. 
Bayesian CL \cite{li2025became, kumar2021bayesian, bonnet2025bayesian,thapabayesian} maintains parameter uncertainty via approximate posteriors, but typically relies on complex inference machinery. Fig. \ref{fig1} compares representative CL strategies on label-permuted EMNIST (80 tasks) and input-permuted CIFAR-10 (200 tasks). Although these methods reduce forgetting compared to naive SGD, their average accuracy still degrades as more tasks are seen, and they incur growing costs in memory, model size, or training complexity  \cite{van2024continual}.
Among these strategies, regularization-based methods \cite{kirkpatrick2017overcoming,shen2024continual, fayek2020progressive, parisi2019continual} are particularly attractive because they operate within a single model without replay buffers or architectural expansion. They preserve prior knowledge by augmenting the task loss with a penalty on deviations from previously learned parameters, $\mathcal{L}_{\text{Total}}\!=\!\mathcal{L}_{\text{New}} + \Omega_{\text{Old}}$, where $\Omega_{\text{Old}}$ is an $\ell_2$-based constraint. When optimized with standard gradient descent (e.g., SGD), this formulation blends learning and retention into a single update direction \cite{lyle2024disentangling, elsayed2024addressing, li2025became}, as illustrated in Fig. \ref{CF}. This gradient blending reduces forgetting but enforces a non-selective trade-off in which all parameters are treated uniformly, limiting the model's ability to adapt even in directions that would not harm prior performance.

Effective CL therefore requires selective knowledge preservation, recognizing that not all parameters need to be constrained equally when learning new tasks. Ideally, the model should be able to decide which parts of the network remain plastic and which become stable. This motivates sparsity-based regularization, such as $\ell_1$, which induces sparse deviations from previously learned solutions. By restricting changes to task-relevant parameters (those contributing to prior tasks), the model can preserve existing knowledge while keeping unused capacity available for future tasks. However, the $\ell_1$ penalty is difficult to handle with standard gradient solvers \cite{ziyin2023spred, wasserman2004all, sun2015feature}. As a result, existing methods often favor smoother $\ell_2$, not because they are optimal, but because it is easier to optimize in practice.  
We address this limitation by adopting an $\ell_1$ retention mechanism that promotes sparse updates: many parameters, particularly those critical for past tasks, remain unchanged, while adaptation is concentrated on a subset of parameters whose updates do not interfere with past knowledge. To make this formulation tractable, we use the Douglas-Rachford Splitting (DRS) \cite{douglas1956numerical,eckstein1992douglas} to decouple task learning from non-smooth regularization. In our setting, DRS is applied to objectives of the form $\min(f(x) + g(x))$, where \(f(x)\) is a task loss and \(g(x)\) is a non-smooth regularizer with a simple proximal operator. By alternating between a plasticity step, which adapts the model to the current task, and a stability step, which enforces sparse retention via the proximal map of $g$, DRS avoids gradient blending and yields selective parameter updates. This guided interaction produces solutions in a less redundant parameter subspace that supports both knowledge retention and forward transfer (see Fig.~\ref{workflw}).
%%%%%--------------------
%
Our model offers three advantages: \circled{1} Rather than relying on replay, architectural expansion, or complex Bayesian approximations, we formulate stability and plasticity as two interacting operators computed via an operator-splitting framework. \circled{2} DRS enables $\ell_1$-based selective knowledge retention, which is difficult to realize with standard gradient descent. \circled{3} By decoupling learning and preservation, the stability step shapes the parameter space in a way that facilitates adaptation to new tasks, yielding continual learners that maintain stability while promoting forward transfer.

%%%%%%%%%%%%%%%%%%%%%%%%
%%%%%%%%%%%%%%%%%%%%%%%%

\section{Related Work}

\subsection{The Need for Continual Learning}
%Continual learning (CL) 
CL enables AI to learn from a stream of tasks/skills over time, accumulating new knowledge without forgetting what was learned previously \cite{french1999catastrophic, polson2015statistical}. In the open world, this adaptability offers substantial benefits, e.g., a medical diagnostic tool must identify novel virus strains without losing the ability to detect common illnesses, or a service robot should add new cleaning routines to its repertoire without unlearning how to navigate a room \cite{dohare2024loss, van2024continual}. However, standard neural networks struggle with this due to catastrophic forgetting: learning new tasks often overwrites parameters important for past ones. This creates a difficult trade-off between plasticity (learning new things) and stability (remembering old things) \cite{mermillod2013stability}. Existing approaches address this dilemma through varying strategies. 

\noindent{\bf{Stability-centric methods: }} They aim to prevent forgetting, either by freezing parameters or replaying past data. Parameter isolation methods, such as dynamic architecture expansion, freeze parts of the model or allocate task-specific parameters \cite{feng2022progressive, yoon2017lifelong,  saha2023continual, kang2022forget, malviya2022tag, cai2025rehearsal}. While effective in retention, they scale poorly as memory and compute grow with the number of tasks. Rehearsal-based methods store a buffer of past samples or use generative models for replay during training \cite{chaudhry2018efficient, rudner2022continual, hayes2020remind, eskandar2025star}. Though they improve stability, they raise concerns about storage and scalability. Regularization-based methods extend the loss function with penalty terms to restrict updates on important parameters \cite{kirkpatrick2017overcoming,  aljundi2018memory, batten2024tight, thapabayesian}. These models are scalable and replay-free. Innovations such as SFSVI \cite{rudner2022continual} extend this idea to function space regularization. However, most use $\ell_2$ penalties, which act as passive constraints,  preserving all parameters uniformly without encouraging selective retention or generalization \cite{dohare2024loss, elsayed2024addressing}.

\noindent {\bf{Plasticity-centric methods: }} Recent works have highlighted that CL failures are not only due to forgetting but also to a loss of learning capacity over time \cite{dohare2021continual, lyle2024disentangling, batra2024evcl}. Representative solutions include continual neuron reinitialization \cite{dohare2024loss}, network expansion \cite{nikishin2023deep}, activation reshaping \cite{abbas2023loss}, and regularization toward initialization \cite{kumar2021bayesian, eskandar2025star}. Meta-learning and Bayesian posterior updates are also widely used to improve plasticity \cite{batten2024tight, li2025became, wu2024mitigating, lee2024recasting, lee2024approximate}. While these models enable fast adaptation, most of them either (i) sacrifice stability unless combined with replay, or (ii) require task boundaries or heavy architectural changes.

\noindent {\bf{Targeting both stability and plasticity: }} A third line of methods explicitly balances stability and plasticity objectives. For example, UPGD balances the two objectives by weighting parameter updates according to their estimated utility \cite{elsayed2024addressing}. BECAME employs a Bayesian merging strategy to retain informative posterior distributions during adaptation \cite{li2025became}. Other approaches, such as Flashback learning \cite{mahmoodi2025flashbacks}, employ task-targeted rehearsals, while dual-memory systems \cite{kim2023achieving} use separate subnetworks for new and previous tasks, coupled with consolidation mechanisms. Bayesian learners \cite{bonnet2025bayesian, thapabayesian} further leverage uncertainty estimates to balance consolidation with flexibility. While promising, these approaches often add complexity, rely on explicit task boundaries, or require memory-intensive replay buffers \cite{van2024continual}. In contrast, regularization-based methods offer a more scalable and memory-efficient alternative \cite{lyle2024switching,dohare2024loss, ahn2019uncertainty, chen2021overcoming}. By encoding prior knowledge as a penalty term within the loss function, they preserve past information without expanding the model or storing large datasets. However, their standard reliance on dense $\ell_2$ regularization imposes a uniform constraint across all parameters, reinforcing the stability-plasticity trade-off. Because such penalties act globally over the parameter space, they fail to differentiate between task-essential and redundant weights, leading to interference that limits adaptability and forward transfer \cite{polson2015statistical, yoo2024layerwise}. We argue that CL models should instead focus on selective preservation, protecting essential weights while pruning redundant ones to free capacity for future tasks. This motivates our focus on non-smooth regularizers $\ell_1$ \cite{ziyin2023spred}, which is non-differentiable at zero and thus promotes sparsity.

\begin{figure}[t]
    \centering
    \includegraphics[width=0.88\linewidth]{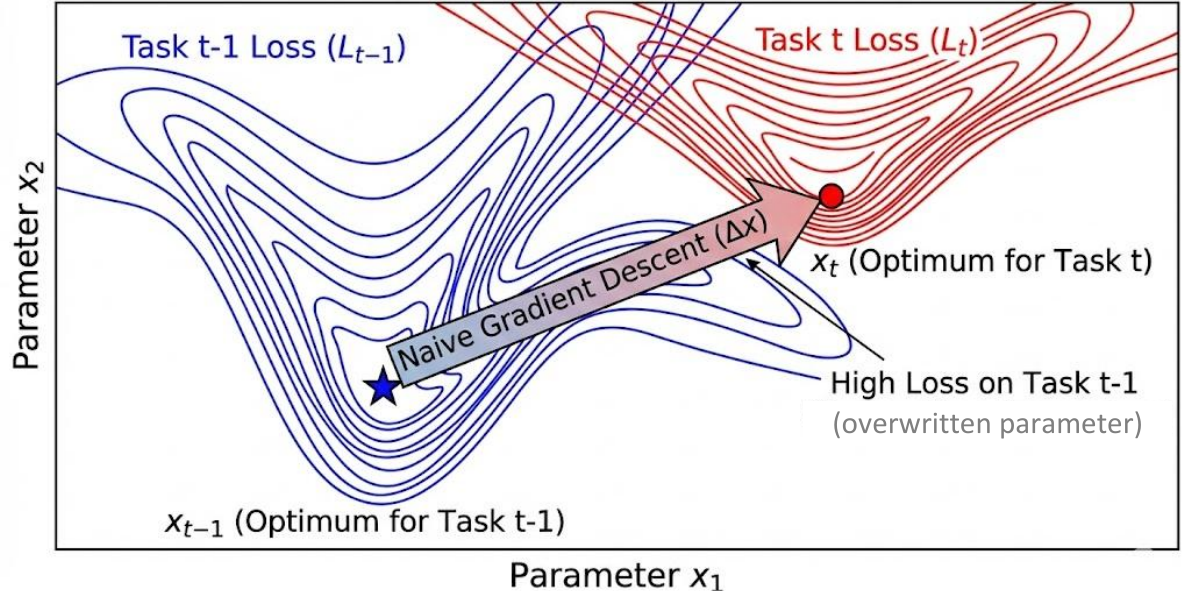}
    \caption{Catastrophic forgetting in CL can be understood as an optimization conflict: gradients computed for the current task $t$ often point in directions that increase the loss of previously learned tasks $t-1$. }
    \label{CF}
\end{figure}

\vspace{-5pt}

\subsection{Operator Splitting Solvers}
Douglas-Rachford Splitting (DRS) \cite{douglas1956numerical, gabay1976dual} is a classical operator-splitting algorithm for minimizing composite objectives $\min_x f(x) + g(x)$, where \(f\) and \(g\) are two distinct functions. In our setting, this formulation represents the two key objectives of CL: $f$ captures plasticity (task learning) and $g$ enforces stability (knowledge retention). Unlike gradient-based training, DRS does not merge these objectives into a single loss. Instead, it allows each to compute independently and then reconciles them through a consensus step. This redefines CL as a negotiation between two objectives rather than a compromise within one gradient. A key advantage of DRS is its ability to handle non-smooth or constrained functions via proximal operators \cite{aljadaany2019douglas, stellato2020osqp, garstka2021cosmo, mai2022fast}. This makes it particularly suitable for problems involving non-differentiable regularizers such as the $\ell_1$ \cite{ziyin2023spred} or indicator functions, where standard gradient-based solvers are not applicable \cite{wasserman2004all, sun2015feature}. Because of these strengths, DRS has been adopted across domains such as signal processing, control, and machine learning \cite{tran2021feddr, mai2022fast, stellato2020osqp, li2016douglas, zareapoor2024rethinking, anshika2024three, ozaslan2025accelerated}, where it enables modular optimization of data-fitting and regularization components within a unified framework. Some prior work has explored the decoupling strategy in CL objective (\cite{yoo2024layerwise, polson2015statistical, zhao2023rethinking}), within replay or Bayesian settings. In contrast, we employ DRS to decouple the two objectives of CL, interpreting $f$ as the task-learning term and $g$ as the stability-preserving regularizer. This is achieved without relying on replay buffers or architectural modifications.

\subsection{Decoupled vs. Fused Optimization}
\begin{figure*}[t]
    \centering
    \includegraphics[width=0.88\textwidth]{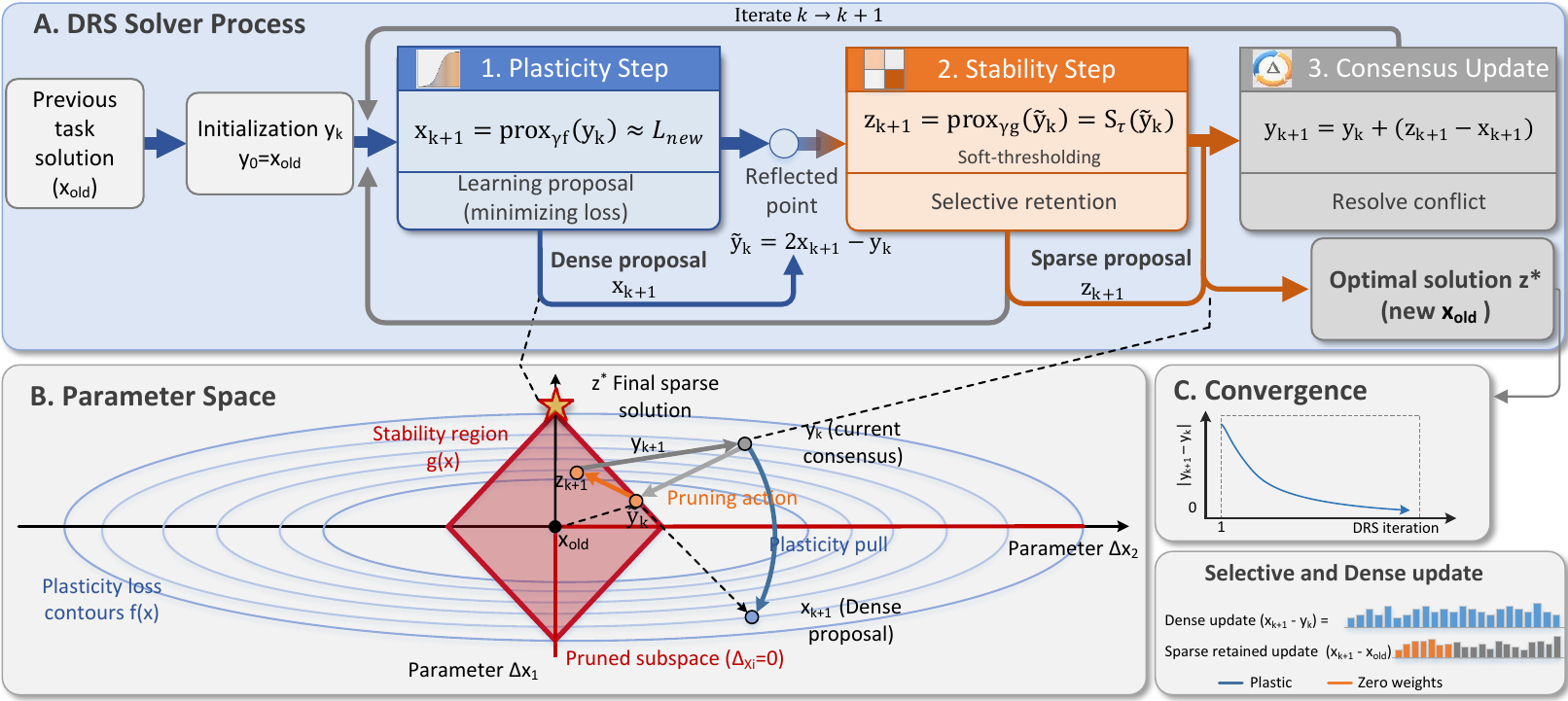} \vspace{-8pt}
    \caption{Illustration of the continual learner based on DRS in parameter update space (where $\Delta x_1, \Delta x_2$ denote representative update directions). Learning is decomposed into a plasticity step $f$ (blue), which generates a dense proposal by minimizing the task loss $\mathcal{L}_{new}$, and a stability step $g$ (orange), which enforces knowledge retention via a non-smooth proximal operator. At the start of each task, the solver is initialized at the previous solution $y_0=x_{old}$;  the plasticity step $x_{k+1}$ is reflected ($\tilde{y}_k$) and passed to the stability step, where an $\ell_1$ proximal mapping applies soft-thresholding to induce sparsity in $\Delta x$ by setting some components to zero (task-essential) while allowing others to change ($\Delta x_2 \neq 0$). The consensus variable is updated ($y_{k} \to y_{k+1}$), to resolve conflicts between the competing objective until the solver converges to an optimal solution  $z^\star$ that is used as the parameter state for the next task. Our proposed model is detailed in Algorithm~\ref{alg} and Sec. \ref{discuss}. }
    \label{arch}
\end{figure*}

Our framework differs from existing continual-learning approaches in both its optimization strategy and preservation mechanism (see Fig. \ref{arch}). Unlike prior work like, SGP \cite{saha2023continual}, UPGD \cite{elsayed2024addressing}, UCL \cite{ahn2019uncertainty}, BAN \cite{thapabayesian}, POCL \cite{wu2024mitigating}, MESU \cite{bonnet2025bayesian}, and SB-MCL \cite{lee2024learning}, which fuse task and stability objectives into a single loss, our model uses a DRS solver to decouple them. This formulation negotiates between plasticity and stability, reaching a consensus solution that satisfies both goals, while mitigating gradient interference. In addition, while recent works such as BECAME \cite{li2025became}, CAI \cite{cai2025rehearsal}, and STAR \cite{eskandar2025star} encourage sparsity through Bayesian pruning, masking, or replay buffers, our model embeds sparse selectivity directly into the optimizer via an $\ell_1$ proximal operator. This mechanism enables the solver to identify and preserve a minimal subset of parameters through the solver's iterations.

%%%%%%%%%%%%%%%%%%%%%%%%%%%%%
%%%%%%%%%%%%%%%%%%%%%%%%%%%%%

%%%%%%%%%%%%%%%%%%%%%%%%%%%
%%%%%%%%%%%%%%%%%%%%%%%%%%%

\section{Our Proposed Approach}

\subsection{Optimization Challenges in Continual Learning}\label{problem}
In continual learning, the objective is to minimize the cumulative loss over a sequence of tasks $1,\dots,T$: $\min_x\sum_{t=1}^T\mathcal{L}_t(x)$. However, in many realistic settings, data from previous tasks ($D_{1:t-1}$) is unavailable when learning task $t$, making it infeasible to compute the total loss directly. A common workaround is to regularize the current task loss 
\begin{equation}
\min_x \; \mathcal{L}_t(x) + \lambda \ \Omega(x; x_{t-1}),
\end{equation} 
where $x_{t-1}$ represents the parameters learned up to the previous task. Fig. \ref{CF} illustrates how minimizing $\mathcal{L}_t$ alone can degrade performance on $\mathcal{L}_{t-1}$. Despite the variety of existing penalties $\Omega$,  \cite{elsayed2024addressing, wu2024mitigating, kirkpatrick2017overcoming, lyle2024disentangling}, this surrogate objective faces two limitations. {\it1) Lack of parameter selectivity}: most methods use an $\ell_2$-based penalty, where $\Omega \propto \|x - x_{t-1}\|^2_2$, and $\nabla\Omega(x)=2(x-x_{t-1})$. This penalty applies a nonzero update to every parameter where $x_i \neq (x_{t-1})_i$, regardless of its importance to past tasks \cite{ziyin2023spred}. When computing the totall update via gradient descent
\begin{equation}
\Delta x \propto -(\nabla \mathcal{L}_t(x) + \lambda \nabla \Omega(x)),
\label{update}
\end{equation}
the learning signal ($\nabla \mathcal{L}_t$) and the retention constraint ($\nabla \Omega$) are combined into a single update vector \cite{li2025became, dohare2024loss}. This results in global, non-selective updates, limiting the model's ability to preserve important parameters from previous tasks while adapting others to new tasks. {\it2) Incompatibility between sparse regularizer and standard solvers}:  Ideally, continual learning should permit only a subset of parameters to change per task, while preserving the rest ($\Delta x_i\!=\!0$). Sparse regularizers (i.e., $\ell_1$) can encourage this behavior \cite{wasserman2004all, ziyin2023spred}: $\Omega = \|F \odot (x - x_{t-1})\|_1$, where $F$ is an importance weight matrix and $\odot$ denotes elementwise multiplication. Unlike $\ell_2$, the $\ell_1$ norm encourages exact zeros in parameter updates, supporting selective learning \cite{ziyin2023spred}. However, $\ell_1$ is non-differentiable at zero and thus incompatible with standard gradient solvers like SGD \cite{sun2015feature, ziyin2023spred}. While subgradient and proximal methods exist, they are rarely applied in deep continual learning due to computational challenges \cite{polson2015statistical, hertrich2023proximal}. These limitations highlight a mismatch between the optimization methods used in CL and the regularizers best suited for selective learning. To address this gap, we adopt an optimization strategy based on the DRS algorithm, a proximal method designed to minimize composite objectives involving both task loss and non-smooth regularization. Although classical DRS requires exact proximal operators (intractable for deep networks), we follow \cite{aragon2020douglas, tran2021feddr, li2016douglas} that apply approximate variants in deep learning settings. This yields a tractable solver that avoids gradient blending and enables selective parameter updates (see Fig.~\ref{arch}).

\vspace{-5pt}

\subsection{DRS-based Continual Learner}
\begin{figure}[H]
    \centering
    \includegraphics[width=0.51\textwidth]{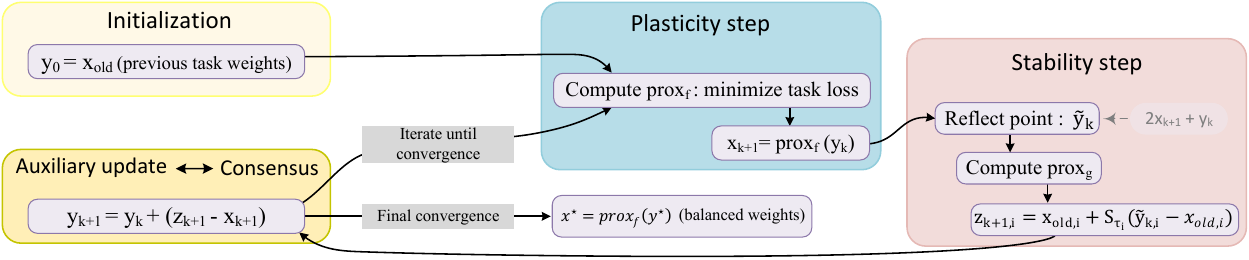}
    \caption{Processing steps of the DRS continual learner.}
    \label{workflw}
\end{figure}

To address the lack of parameter selectivity in continual learning, we propose a proximal update scheme inspired by DRS algorithm. Specifically, our formulation decouples task learning (\(f\), plasticity) and knowledge retention (\(g\), stability) into two distinct components. We model continual learning as a negotiation between two components. For each task \(t\), the learner minimizes the following objective 
\begin{equation}
 \min_{x} \{\underbrace{\mathcal{L}_{t}({x}; D_t)}_{\text{plasticity}, f({x})} + \lambda \cdot \underbrace{\sum_{i} F_i |x_i - x_{\text{old}, i}|}_{\text{stability}, g({x})} \},  
  \label{main}
\end{equation}
where \(x_{\text{old}}\) are the parameters from the previous task $t-1$, and \(F_i > 0\) represents the importance weight (e.g., diagonal Fisher information computed after task \(t-1\)). The \(\ell_1\) 
penalty on parameter differences encourages sparse changes, implicitly partitioning the parameters into
\begin{equation}
\begin{aligned}
\mathcal{S}=\{i:x_i=x_{\text{old},i}\} \quad \text{(stable parameters)}, \\
\mathcal{P} = \{ i : x_i \neq x_{\text{old},i} \} \quad \text{(plastic parameters)}.
\end{aligned}
\end{equation}
since the regularizer $g(x)$ is non-smooth, standard gradient descent struggles with this composite structure. Instead, we adopt approximate DRS and introduce an auxiliary variable \(y_k\) (consensus anchor), initialized as \(y_0 = x_{\text{old}}\). At each iteration \(k = 0, 1, \dots\), we perform the following three steps (also shown in Fig. \ref{workflw}).

\vspace{-5pt}

\subsubsection{Plasticity Proposal}\label{step1}
We first generate a task-specific proposal $x_{k+1}$ that focuses on learning the current task. Here, $y_k$ serves as the input anchor for both learning and stability
\begin{equation}
  x_{k+1} = \text{prox}_{\gamma f}(y_k)=\arg\min_{x} \ \{ \mathcal{L}_{\text{t}}(x) + \frac{1}{2\gamma}\|x - y_k\|^2\}  
  \label{plas}
\end{equation}
This step seeks a parameter vector that improves performance on the current task while remaining close to $y_k$. \\
{\bf{Computation:}} Solving (\ref{plas}) exactly is intractable for deep networks. We therefore approximate the proximal step with a gradient descent update \cite{stellato2020osqp, ozaslan2025accelerated}, $x_{k+1}\!\approx\!{y}_k-\eta \cdot \nabla \mathcal{L}_{t}({y}_k, D_t)$, where $\eta$ is the step size (we set $\eta=\gamma$). This step allows the model to adapt to task $t$, but produces dense updates, affecting all parameters regardless of their relevance to prior tasks. In the next step, this candidate is refined by the stability mechanism.

\begin{figure}[H]
    \centering
    \includegraphics[width=0.95\linewidth]{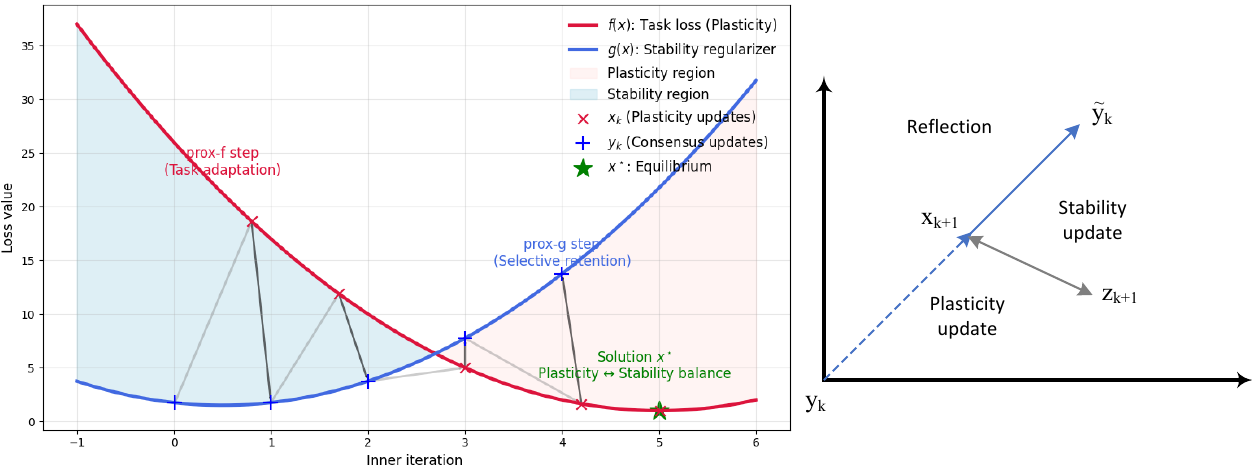}
    \caption{Alternating proximal operators between the task loss and regularizer. The consensus variable $y_k$ is updated iteratively to balance learning and retention, converging to a stable solution. On right, the reflected point $\tilde{y}_k$ amplifies the learning proposal $x_{k+1}$, creating a parameter-space deviation where the stability update $z_{k+1}$ can selectively preserve or prune weights. This reflection enables the stability mechanism to identify which parameters conflict with prior knowledge, guiding the model toward an optimal solution.}
    \label{reflection}
\end{figure}

\subsubsection{Stability Filter}\label{step2}
To preserve knowledge from earlier tasks, we apply the exact proximal operator of the sparsity-inducing regularizer \(g(x)\) to the reflected point $\tilde{y}_k = 2x_{k+1} - y_k$, which captures the deviation introduced by the plasticity 
\begin{equation}
z_{k+1} = \text{prox}_{\gamma g}(\tilde{y}_k),
\end{equation}
where $g(x)=\sum_i F_i |x_i-x_{\text{old},i}|$. \\
{\bf{Computation:}} Since $g(x)$ is a positive weighted $\ell_1$ norm, this proximal operator has a closed-form solution based on soft-thresholding \cite{ziyin2023spred, sun2015feature}, which is computed as
\begin{equation}
z_{k+1,i} = x_{\text{old},i} + S_{\tau_i} ( \tilde{y}_{k,i} - x_{\text{old},i})  \ \  \text{with} \ \tau_i = \gamma \lambda F_i 
\label{threshold}
\end{equation}
where $S_{\tau}(u)=\text{sign}(u)\cdot \max(|u|-\tau, 0)$. This operator suppresses small, unimportant updates (based on $F_i$) while allowing significant ones to proceed ($\Delta_i=\tilde y_{k,i}-x_{\text{old},i}$), partitioning parameters into stable and plastic subsets
\begin{itemize}
\item {Retention: } If $|\Delta_i|<\tau_i$, then $z_{k+1,i}=x_{\text{old},i}$. The proposed change is rejected and the parameter remains fixed at its previous value.
\item {Adaptation: } If $|\Delta_i|\ge\tau_i$, then $z_{k+1,i}\neq x_{\text{old},i}$. The task pressure is high enough to justify a change, and the parameter is allowed to adapt to the current task.
\end{itemize}
Indeed, the stability step filters the dense task-driven proposal $x_{k+1}$, ensuring that only important updates (relative to importance $F_i$) are retained. This ensures that adaptation occurs in regions where it is most beneficial for the current task and least damaging to prior. 
%%%%
The regularizer $g(x)$ is closed and convex, as it is a positive weighted $\ell_1$ norm \cite{aragon2020douglas, garstka2021cosmo}, also shown in propositions \ref{l1}\ - \ \ref{convex}). %

\subsubsection{Consensus Update}\label{step3}
After computing the plasticity proposal and the stability filter we update the auxiliary variable $y$ via a correction step 
\begin{equation}
{y}_{k+1} = {y}_k + \underbrace{({z}_{k+1} - {x}_{k+1})}_{\text{correction vector}}
\label{final}
\end{equation}
%%%
This step propagates the disagreement between the plasticity and stability components into the next iteration. When the two updates agree (i.e., $z_{k+1} \approx x_{k+1}$), the consensus stabilizes and $y_{k+1} \approx y_k$.  Otherwise ($z_{k+1}\neq x_{k+1}$), the correction vector stores the conflict between plasticity and stability, and the next anchor $y_{k+1}$ is shifted to resolve that conflict. 
This process generates a sequence (${y_0, y_1, \ldots, y_k}$) that continues until the conflict is minimized ($\|y_{k+1} - y_k\|\approx0$). While standard DRS convergence is proven under convexity and exact proximal updates, our setting follows \cite{aragon2020douglas, ozaslan2025accelerated, tran2021feddr, li2016douglas} and approximate gradient-based proximal step, ensuring the final solution $x^\star$ is a verified intersection between stability and plasticity (see Theorem \ref{theory}).

\paragraph{Discussion.}  This procedure separates task adaptation from memory preservation by decoupling the updates for $\mathcal{L}_t$ and $\Omega$. It avoids gradient blending and enables parameter-level selectivity, which is key challenges in continual learning. We note that our approach is an approximate version of DRS, following \cite{aragon2020douglas, li2016douglas, stellato2020osqp}.

\begin{table}[h]
\centering
\caption{Summary of notation.}\label{notion}
\scalebox{0.85}{
\begin{tabular}{ll}
\midrule
\textbf{Symbol} & \textbf{Description} \\
\midrule
$T$& task index  (1 to $T$) \\
$D_t$& dataset for task $t$ (with input-label pairs $\{(u_i, v_i)\}$) \\
$x\in\mathbb{R}^d$& neural network parameters (model weights) \\
$\mathcal{L}_t$& task-specific loss function (plasticity objective)\\
$S_{\text{old}}$& subspace of previous tasks ($1 \dots t-1$) \\
$k$ & DRS iteration index (1 to $K$) \\
$x^{(k)}$& plasticity/learning variable at solver iteration $k$ \\
$z^{(k)}$& stability variable, which is the output of the proximal step  \\
$y^{(k)}$& anchor point between the split steps, tracking the gap ($x-z$) \\
$\lambda$& regularization strength (for $\ell_1$ penalty)   \\
$\gamma$& step size for the splitting algorithm   \\
$\Delta x$& update vector ($x_t-x_{t-1}$) constrained to the null space \\
$\text{prox}_{\gamma f}$ & proximal operator associated with function $f$  \\
$\|\cdot\|_1$& $\ell_1$-norm used to induce sparsity  \\
$\nabla \mathcal{L}$& stochastic gradient of the loss function \\
$\langle\cdot,\cdot\rangle$& inner product operator \\
\bottomrule
\end{tabular}}
\end{table}

\vspace{-7pt}

\subsubsection{Task-to-Task Transition}
The sequence  $y_0 \to y_1 \to \dots \to y_k$ integrates both forces, converges to a fixed point $y^\star$ that balances task performance and stability constraints. Importantly, to ensure the model benefits from selective retention, we adopt the output of the stability step, $z^\star$, as the final model parameters for task $t$. This guarantees that the sparsity constraints are respected, i.e., parameters important to prior tasks remain unchanged (\( x_i = x_{\text{old},i} \) for all \( i \in \mathcal{S} \)).  As a result, unnecessary updates are pruned, and the model remains in a compact subspace that preserves unused capacity for future learning. Once task $t$ is complete, the retained parameters \(z^\star\) become the \( x_{\text{old}} \) for task \(t+1\), and the importance weights \( F \) are recomputed based on the updated model.

\begin{algorithm}[t] \small
\begin{algorithmic}
\caption{Our proposed DRCL. A summary of the key notations used throughout this work is provided in Table \ref{notion}.} \label{alg}
\Require current task data $T$, inner iterations $I_{\max}$, importance $F$, learning rate $\eta$, sparsity weight $\lambda$, past  parameters $x_{\text{old}}$.
\For{each task/dataset in $\mathcal{T}$}
\State Compute Fisher information $F_i$ based on $x_{\text{old}}$ 
\State Set consensus (anchor point) $y_0 \leftarrow x_{\text{old}}$
\For{iteration $k = 0$ $\rightarrow$ MaxIter}
\State \textbf{1. Plasticity pull ($\text{prox}_{f}$)}
\State \quad Compute gradient $g_k = \nabla \mathcal{L}_t(y_k; \mathcal{D}_t)$
    \State \quad $x_{k+1} \leftarrow y_k - \gamma \cdot g_k$  \quad (learning proposal)
\State \textbf{2a. Reflection }
    \State \quad $\tilde{y}_k \leftarrow 2x_{k+1} - y_k$
%%\State \; $\tilde{y}_i \leftarrow 2x_{i+1}-y_i$
  %  
    \State \textbf{2b. Stability $\text{prox}_{g}$ (pruning)}
\State \quad Compute thresholds $\tau = \gamma \lambda F$
    \State \quad $z_{k+1} \leftarrow x_{\text{old}} + \text{soft\_threshold}(\tilde{y}_k - x_{\text{old}}, \tau)$
    \State \textbf{3. Consensus update}
    \State \quad $y_{k+1} \leftarrow y_k + (z_{k+1} - x_{k+1})$
\EndFor
\EndFor
\Return $z^\star$ (Final Sparse Weights)
\end{algorithmic}
\end{algorithm}

%%%---------------------
%%-------------

\begin{theorem}\label{theory}
At convergence, the solution $z^\star$ is guaranteed to satisfy the stability condition $|z_i^\star - x_{\text{old},i}| = 0$ for all parameters $i$ where the difference between the reflected anchor and the previous parameters is smaller than the importance threshold: $|2x_{i,k+1} - y_{i,k} - x_{\text{old},i}| < \tau_i \implies z_{i,k+1} = x_{\text{old},i}$. 
\end{theorem}

\noindent This result shows that if the plasticity proposal does not push a parameter beyond its importance weight (which is defined by $\tau_i = \gamma \lambda F_i$), then the stability operator will return that parameter to its previous value. Compared to standard proximal gradient methods \cite{anshika2024three, aragon2020douglas, garstka2021cosmo}, which apply soft-thresholding directly to the gradient step, our DRS formulation uses a reflected point in step 2, $2x - y=x+(x-y)$, that amplifies disagreement over time.  If a proposed update conflicts with the stability constraint, the correction term accumulates in the auxiliary variable $y$, increasing the likelihood that the soft-thresholding operator will reject the update. This mechanism ensures that the optimization converges only when plasticity updates do not interfere with parameters critical to previous tasks. Consequently, the final solution $z^\star$ achieves sparsity in its updates, enabling task adaptation while preserving prior knowledge.

\begin{proposition}\label{l1}
The $\ell_1$ penalty enforces structural sparsity in parameter updates, where non-essential changes ($\Delta x_i$) are set to zero. This property is necessary to maximize plasticity by preserving the network available capacity.   
\end{proposition}

\noindent The stability objective $g(x)$ is a $\ell_1$ norm, which is a closed and convex function \cite{ozaslan2025accelerated, aragon2020douglas, stellato2020osqp}. The optimality condition for the objective is given by  $0\in \nabla_i{f}(x^\star) + \partial{g}_i(x^\star)$, where $\partial g_i$ is the subdifferential. For the non-smooth $\ell_1$, the gradient at the point of zero change ($\Delta x_i = 0$) is the interval $[-\lambda F_i, \lambda F_i]$. Consequently, a parameter remains unchanged from its previous state ($x_i^\star = x_{\text{old},i}$) if $|\nabla_i f(x^\star)| \le \lambda F_i$. 
This condition demonstrates that the optimal solution $x^\star$ achieves sparsity ($\Delta x_i = 0$) even when the task-loss gradient $\nabla_i f$ is non-zero. This mechanism enables selective retention: parameters where the task-learning pressure is lower than the importance threshold ($|\nabla_i f| \le \lambda F_i$) are kept stable, while adaptation is restricted to parameters where the gradient exceeds the regularization strength. In contrast, smooth $\ell_2$ lack this property; the optimality condition for $\ell_2$ requires $\Delta x_i$ to be non-zero whenever $\nabla_i f$ is non-zero, leading to dense updates that overwrite prior knowledge across the entire parameter space.

\begin{proposition}\label{convex}
The stability is computationally efficient because the proximal operator of the $\ell_1$ objective, $\text{prox}_{\gamma g}$, admits a, closed-form solution. This eliminates the requirement for gradient-based optimization on non-smooth properties.
\end{proposition}
\noindent The proximal operator $\text{prox}_{\gamma g}(y)$ requires solving a minimization that is separable across each parameter dimension $i$: $\text{prox}_{t g}(y)_i = \arg\min_{x_i}\{ \lambda F_i |x_i - x_{\text{old}, i}| + \frac{1}{2t}(x_i-y_i)^2\}$. By performing a change of variables $u_i = x_i - x_{\text{old}, i}$ and defining the threshold $\tau_i = \gamma \lambda F_i$, the objective takes the form of the canonical soft operator $S_{\tau_i}$ \cite{ziyin2023spred}. The optimal solution is thus given by $x_i^\star=x_{\text{old}, i} + S_{\tau_i}(y_i - x_{\text{old}, i})$. The existence of this exact solution ensures that the stability component of the DRS solver is calculated via simple algebraic operations. This resolves the inherent challenge of non-differentiability that prevents the direct integration of $\ell_1$ penalties into standard gradient descent pipelines.

\begin{table*}[t]
\centering
\begin{tabular}{@{}l|lccc>{\columncolor{orange!15}}ccc>{\columncolor{orange!15}}c@{}}  \toprule
& \multicolumn{5}{c|}{Disjoint Tasks (class-split)} & \multicolumn{3}{c}{Joint Tasks} \\ \cmidrule(lr){2-6}\cmidrule(lr){7-9} 
Method & {CIFAR-100 [10]} & {CIFAR-100 [20]} & {TinyImgNet [20]} & {ImgNet [100]} & {Avg.} & {CelebA [20]} & {EMNIST [20]} & {Avg.} \\ \midrule
Multi-task & \(76.1\pm0.3\) & \(79.3\pm0.5\) & \(60.5\pm1.2\) & \(63.9\pm0.5\) & 69.9& \(88.2\pm0.2\) & \(87.9\pm0.8\) & 88.0 \\   
Baseline & \(54.9\pm2.1\) & \(56.2\pm0.7\) & \(42.2\pm0.8\) & \(34.7\pm0.9\) & 46.5 &  \(84.6\pm1.2\) & \(86.5\pm0.4\) & 85.6 \\  \midrule 
%EVCL & $66.4\pm0.9$ & $74.7\pm0.6$ & $44.6\pm0.9$ & $45.3\pm1.0$ &57.6 & $81.5\pm1.5$ & $86.2\pm0.3$ & 84.1 \\
A-GEM& $54.4 \pm 1.0$ & $57.8 \pm3.5$ & $42.7 \pm1.2$ & $35.2 \pm1.1$ &47.5 & $83.7\pm1.6$& $87.1\pm0.5$ & 85.4 \\
EWC& $63.5\pm0.8$ & $62.5\pm1.7$ &$42.9\pm0.9$ & $29.6\pm0.9$ & 49.4 & $86.3\pm0.9$ & $86.9\pm0.6$ & 86.6  \\
BAN& {$71.6\pm0.5$} & $\bf{78.9\pm0.8}$ & $58.5\pm0.4$ & $56.0\pm0.5$ & 66.5& {$86.5 \pm 0.7$} & {$87.3\pm0.2$} & 87.1 \\
SB-MCL& ${\bf 72.3\pm 0.3}$ & $78.1\pm0.7$ & {$58.2\pm0.5$} & $56.5\pm0.6$ & 66.3 & $86.9\pm0.5$ & {$87.5\pm0.3$} & {87.0}\\
HAT& $63.9\pm0.5$ & $73.1\pm0.8$ & $53.2\pm1.6$ & $47.6 \pm 1.5$ & 59.4 & $82.5\pm0.6$ & $84.6\pm0.8$ & 83.5 \\
SGP& $72.0\pm0.3$ & $77.8\pm0.6$ & $59.4\pm0.3$ &$57.8\pm0.5$ & 66.7 & $87.3\pm0.2$ & $87.9\pm0.5$ & 87.7 \\
IBPCL& $69.3\pm0.9$ & $76.3 \pm 0.7$ & $57.1\pm 0.6$ & $54.8\pm0.8$ & 64.3 &$87.1\pm 0.3$ & $87.8\pm0.4$ & 87.5 \\
UCL& $65.7 \pm 0.7$ & $74.9 \pm 0.5$ & $55.2 \pm 0.5$ & $40.0 \pm 0.7$ & 58.9 & $86.3\pm0.5$ & $85.2\pm1.2$ & 85.7 \\ 
UPGD&  $71.6\pm 0.2$ &  $78.0\pm 0.9$ & $59.7\pm 0.4$ &  $58.2\pm 0.5$  & 66.9 & $87.0\pm 0.3$ & $87.6\pm 0.3$ & 87.3 \\
POCL& $71.2 \pm 0.4$ & $77.9 \pm 0.8$ & $58.2 \pm 0.6$ & $56.7\pm0.6$& 66.0 &  \(85.7\pm0.9\) & \(86.7\pm0.4\) & 86.2\\
TAG& $62.4 \pm 0.6$ & $69.3\pm 0.7$ & $49.6 \pm 0.5$ & $45.8 \pm 0.3$ & 56.8 & $78.3\pm1.4$ & \(83.9\pm0.5\) & 81.1 \\
WSN& $70.2\pm 0.2$ & $78.2\pm0.6$ & $58.5\pm 0.5$ & $53.2\pm 0.4$ & 65.0 & $84.6\pm0.7$ & $86.0\pm0.2$ & 85.3 \\  
\bottomrule %\rowcolor{lime!20}
Ours & \(71.9\pm0.2\) & $78.5\pm0.6$ & ${\bf61.3\pm0.4}$ & ${\bf60.4\pm0.5}$ & \bf68.0 & $\bf87.8\pm0.3$ & $\bf88.5\pm0.2$ & \bf88.2 \\
\bottomrule
\end{tabular}
\caption{Accuracy (\%) on task-incremental setting for disjoint (split class) and joint (multi-attribute/domains) tasks. Values represent mean $\pm$ standard deviation over 3 runs. Baseline denotes a naive model that learns a new task without considering the previous tasks, and multi-task trains all the tasks together. Our model outperforms regularization and projection baselines across most settings. {\bf{Bold}} text represents the best results.}
\label{alexnet}
\end{table*}

\vspace{-5pt}

\subsection{DRCL Implementation} \label{discuss}
The core implementation of our model reformulates the stability-plasticity as a decoupled optimization problem. Exiting regularization methods sum gradients ($\nabla \mathcal{L} + \lambda \nabla \Omega$), which causes direct vector interference. In contrast, our model decomposes these forces into independent proximal operations. The task-specific adaptation occurs in step 1, while the structural preservation is enforced via the proximal operator in step 2. The consensus step then reconciles these outputs without the interference associated with scalarized objective functions. 
This decomposition enables the uses of $\ell_1$ for selective parameter preservation. Each DRS iteration follows the sequence: $y_k \to x_{k+1}$ (plasticity proposal) $\to \tilde{y}_k$ (reflection) $\to z_{k+1}$ (stability filtering) $\to y_{k+1}$ (auxiliary correction). For each task, we initialize the consensus anchor $y_0 = x_{\text{old}}$. To enable selective retention, we propose the reflected point $\tilde{y}_k = 2x_{k+1} - y_k$, where it doubles the perceived distance between the task proposal and the previous state. This amplification allows the stability operator to distinguish between negligible parameter shifts and critical task adaptations. The filtering result $z_{k+1}$, ensures that small deviations are pruned ($\Delta x_i = 0$) and only essential updates are retained. The discrepancy is then stored in the auxiliary variable $y_{k+1}$ (Eq. \ref{final}), which biases the search space for the subsequent iteration. We observe that setting the $\gamma$ equal to the learning rate $\eta$ ensures convergence within a small number of iterations. At convergence, the final parameters $x^\star$ reside at the intersection of the task-loss and the sparsity-constrained space.

\begin{table}
    \centering
   \scalebox{0.68}{
    \begin{tabular}{@{}l|cccc>{\columncolor{orange!25}}ccc>{\columncolor{orange!25}}c@{}}
      \toprule
        & \multicolumn{5}{c}{Disjoint tasks} & \multicolumn{3}{c}{Joint tasks} \\ \cmidrule(lr){2-6}\cmidrule(lr){7-9} 
       Method & {C-[10]} & {C-[20]} & {TinyINet [20]} & {INet [100]} & {Avg.} & {Celeb [20]} & {EMNIST [20]} & {Avg.}  \\
        \midrule
        A-GEM & -10.5 & -17.6 & -9.8 & -14.2 & -13.1  & -0.4 & +0.6 & +0.1    \\
        EWC & -6.8 & -11.2 & -7.6 & -17.4 &  -10.7  & +1.2 & +0.3 &  +0.7 \\
        BAN & -3.3 & -4.0 & -4.7 & -3.5 & -3.9  & +2.6 & +1.4 & +2.0 \\
        SB-MCL & -3.8 & -4.3 & -4.8 & -3.9 & -4.2  & +2.3 & +0.7 & +1.5 \\
        SGP & -4.2 & -5.1 & -3.6 & -2.4 & -3.8  & +2.9 & +0.8 & +1.9  \\
        IBPCL & -6.4 & -9.5 & -8.6 & -7.9 & -8.1  & +0.9 & +0.3 & +0.6  \\
        UCL & -7.2 & -9.3 & -7.7 & -11.6 & -8.9  & +2.6 & +1.8 & +2.2 \\
        UPGD & -3.1 & -4.6 & -3.7 & -2.7 & -3.5 & +2.5 & +0.9 & +1.7 \\
        POCL & -4.5 & -5.6 & -5.8 & -4.1 &  -5.9 & +1.8 & +0.7 & +1.3 \\
        TAG & \bf-1.2 & \bf-2.8 & \bf-1.4 & -2.0 & \bf-1.8  & +0.3 & +0.1 &  +0.2 \\
           \midrule
        Ours  & -2.8  & -3.2 & -2.5  & \bf-1.7 & -2.5  &  {\bf{+3.7}}  &  \bf+1.9 & \bf+2.8 \\
        \bottomrule
    \end{tabular}
    }
    \caption{Backward transfer result. Lower values indicate forgetting, and positive values indicate improvement due to knowledge transfer. TAG has almost no forgetting, but its forward transfer is limited, resulting in poor final accuracy (Table \ref{alexnet}). Our model achieves the best average knowledge transfer on joint tasks and minimizes forgetting on disjoint tasks. {\bf{Bold}} text represents the best results. }
    \label{backward}
\end{table}

\section{Experiments}
{\bf{Backbone and Datasets: }} We evaluate our model using two backbones: ResNet-18 \cite{he2016deep} and a 5-layer AlexNet \cite{serra2018overcoming}. Following the class-incremental learning (CIL) setting, each task is assigned a separate classifier head. We use standard CL benchmarks including CIFAR-100 \cite{krizhevsky2009learning}, EMNIST \cite{cohen2017emnist}, CelebA \cite{guo2016ms}, TinyImageNet \cite{wu2017tiny} and ImageNet \cite{deng2009imagenet}. These datasets are divided into $T$ sequential tasks; e..e, we evaluate on CIFAR100 10-split (10 tasks, 10 classes each) and 20-Split (20 tasks, 5 classes each). For datasets with low resolution (e.g., $32 \times 32$ for CIFAR-100), we replace the initial $7\times7$ conv (stride 2) and max-pooling with a $3\times3$ conv (stride 1). This modification prevents information loss in the early layers by preserving the spatial dimensions of the feature maps. For the first task ($T_1$), training is similar to the corresponding backbone (with no stability, $\lambda=0$). For subsequent tasks ($T_t, t \ge 2$), the auxiliary variable $y$ is initialized from the final weights of the previous task $y_0^{t} = x_{\text{old}}^{t-1}$. The plasticity step requires one backpropagation pass, while the stability step is an element-wise vector ($O(d)$). Similarly, the consensus update is a simple vector addition of $O(d)$. Consequently, the per-iteration computational overhead is negligible compared to standard SGD.
\begin{figure}
    \centering
    \includegraphics[width=0.47\textwidth]{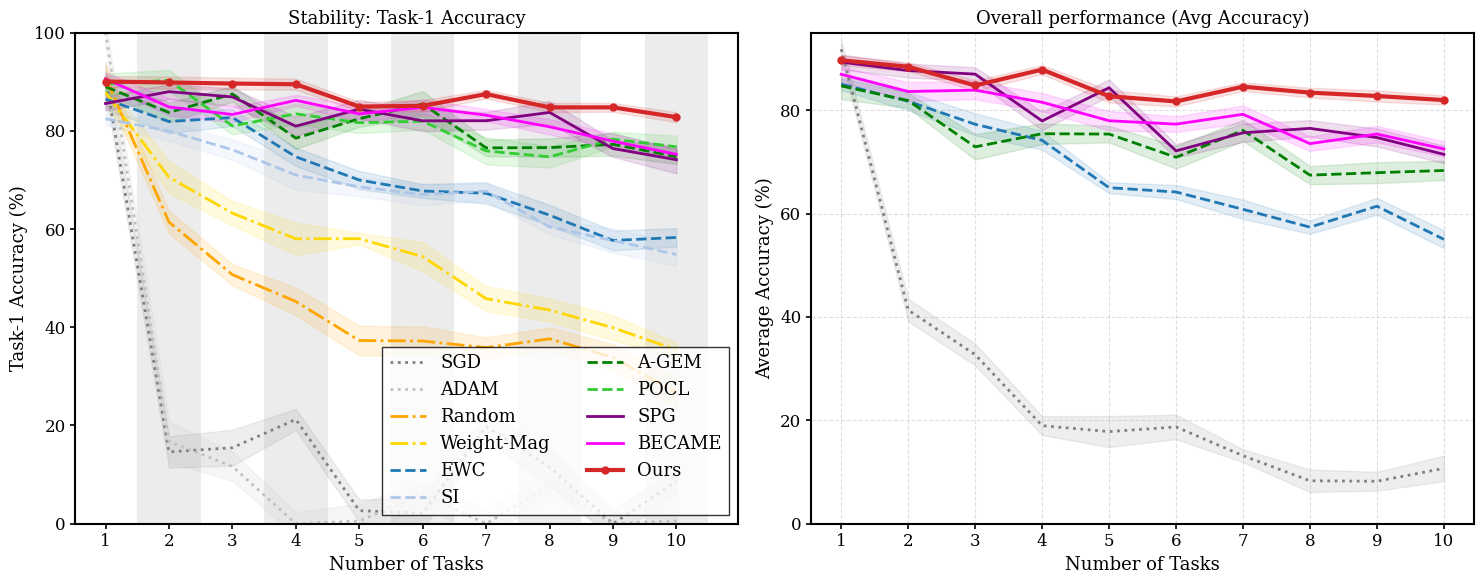}
\caption{Average accuracy shows the plasticity-stability dilemma (right), and accuracy change of the categories learned in the 1st task (left), during sequential learning (Task 1 $\to \dots \to$ Task $10$).}
    \label{plas-stab}
\end{figure}
\begin{figure*}
    \centering
    \includegraphics[width=0.87\linewidth]{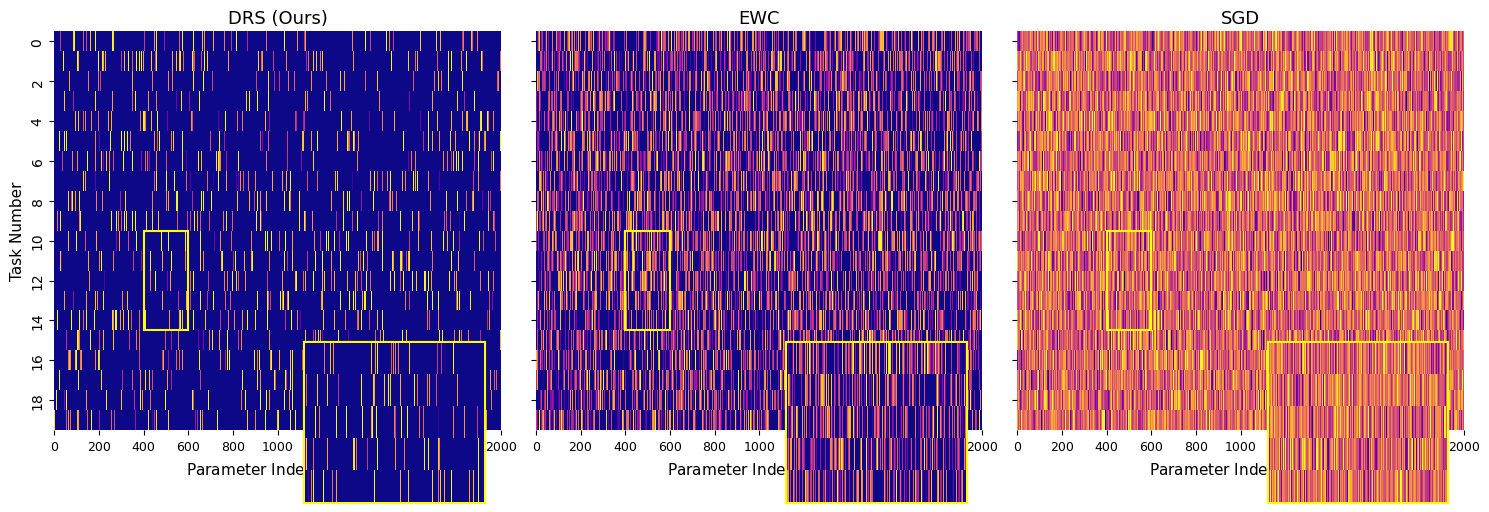}
    \caption{Parameter update magnitudes ($\Delta x$) across 20 tasks. DRS produces sparse, selective updates per task, preserving most parameters (i.e., keeps many updates close to zero, which shows as dark blue regions). By contrast, EWC and SGD update a broader set of parameters per task, indicating weaker or no retention. }
    \label{heat}
\end{figure*}
\indent {\bf{Hyperparameters:}} Following the completion of each task $t-1$, the importance weights $F_i$ are computed using a diagonal approximation of the Fisher Information matrix, averaged over 1,000 random samples \cite{kirkpatrick2017overcoming}. These weights are mean-normalized. We maintain the step size $\gamma$ equal to the learning rate ($\eta\!=\!\gamma$). For the tasks $t \ge 2$ we use $\eta=5\times 10^{-3}$, and $\lambda\!=\!10$. The resulting threshold is $\tau_i=0.05 F_i$, which can effectively filters non-critical parameter updates. We also execute the DRS negotiation for $k=5$ iterations. This process allows the auxiliary variable $y$ to reach a balance between plasticity and stability. Ablation results demonstrate that $k=1$ (single-loop) is insufficient to prevent memory degradation over long sequences.

{\bf{Baselines: }}We use different CL methods as baseline:  EWC \cite{kirkpatrick2017overcoming}, IBPCL \cite{kumar2021bayesian}, A-GEM \cite{chaudhry2018efficient}, SB-MCL \cite{lee2024learning}, UCL \cite{ahn2019uncertainty}, TAG \cite{malviya2022tag}, EVCL \cite{batra2024evcl}, UPGD \cite{elsayed2024addressing}, POCL \cite{wu2024mitigating}, HAT \cite{serra2018overcoming}, BAN \cite{thapabayesian}, SGP \cite{saha2023continual}, BECAME \cite{li2025became},  STAR \cite{eskandar2025star},  MESU \cite{bonnet2025bayesian}, and WSN \cite{kang2022forget}. All results are averaged over 5 random seeds.

\begin{figure*}
    \centering
    \includegraphics[width=0.89\linewidth]{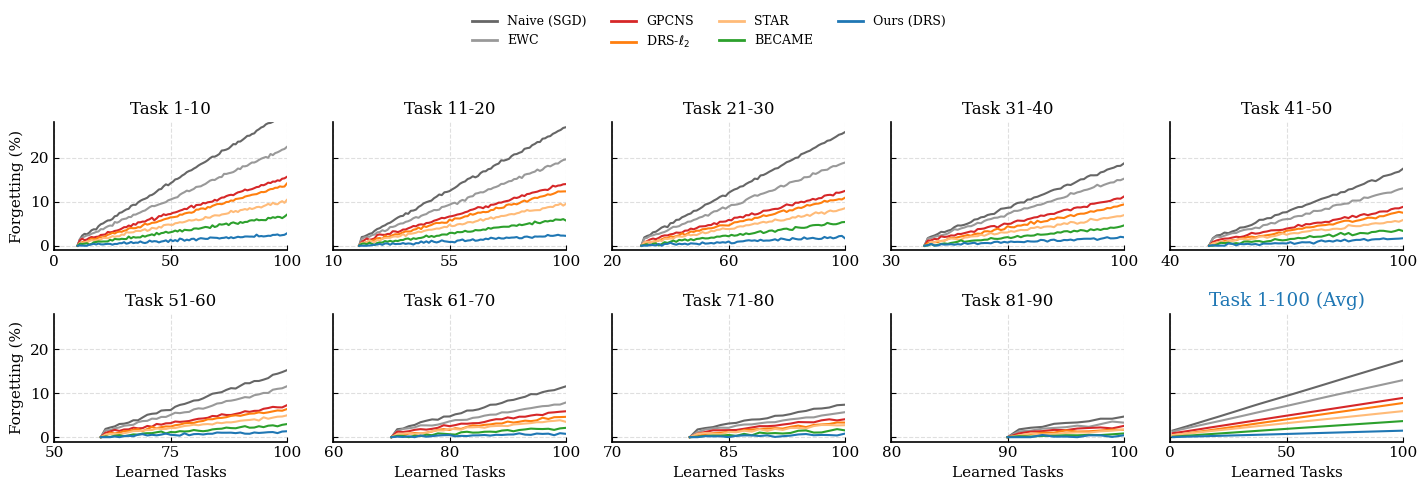}
    %\vspace{-8pt}
    \caption{Granular forgetting over 100 sequential tasks on CASIA-HWDB1.0. The plots illustrate the accumulated forgetting for specific task groups (e.g., tasks 1-10) as the model learns subsequent tasks. The x-axis represents the learning timeline (tasks 0-100) for each group. The average confirms that our model reduces the mean forgetting rate across the entire 100-tasks compared to other CL strategies. }
    \label{forget}
\end{figure*}

\subsection{Comparison with State-of-the-Art Methods}
Table \ref{alexnet} summarizes the performance of different methods on task-incremental learning benchmarks using an AlexNet backbone. We evaluate models on five datasets: CIFAR-100 (10 and 20 splits), Tiny-ImageNet (20 tasks), ImageNet-100 (100 tasks), CelebA (20 binary attributes), and EMNIST (20 tasks).  In the disjoint settings (CIFAR and ImageNet), catastrophic forgetting is the primary challenge due to non-overlapping label spaces. In contrast, the joint task settings (CelebA and EMNIST) share label spaces, emphasizing the importance of knowledge transfer. Average accuracy across all learned tasks is used as the evaluation metric, defined as \(1/T\sum_{t=1}^T \alpha_t^T \), where \(\alpha_i^j\) denotes the test accuracy on task \(i\) after completing task \(j\). The naive baseline, which lacks a forgetting prevention mechanism, performs poorly in disjoint settings  due to parameter interference and feature drift. While regularization methods like EWC improve this, they fail on longer task sequences. Notably, our model outperforms recent high-performing projection methods such as SGP and UPGD. The limitation of projection-based methods lies in their restrictive nature; they block gradient updates that intersect with previous tasks, which can hurt plasticity. Our model, replaces rigid blocking with a decoupled negotiation mechanism. By utilizing the auxiliary variable $y$ and the DRS solver, the model identifies specific parameters that can be modified or shared without compromising existing knowledge. On CIFAR-100, our model remains competitive, though it is slightly outperformed by SB-MCL in the 10-task and BAN in the 20-task. In these lower-resolution, shorter-sequence regimes, Bayesian or Meta-learning priors can provide an advantage in handling uncertainty. However, as task complexity and diversity increase, our proposed strategy becomes the decisive factor. Indeed, meta-learning approaches (SB-MCL) are sensitive to task distributions; when tasks become diverse (as seen in our joint tasks), the meta-prior can become a constraint. The results in Table \ref{backward} provide further insight via backward transfer. While TAG achieves minimal forgetting by locking the model's parameters, this rigidity leads to poor forward transfer, as seen by its lower final accuracy. In contrast, our model achieves comparable stability through iterative DRS negotiation. 
%%%
\begin{figure*}
    \centering
    \includegraphics[width=0.8\linewidth]{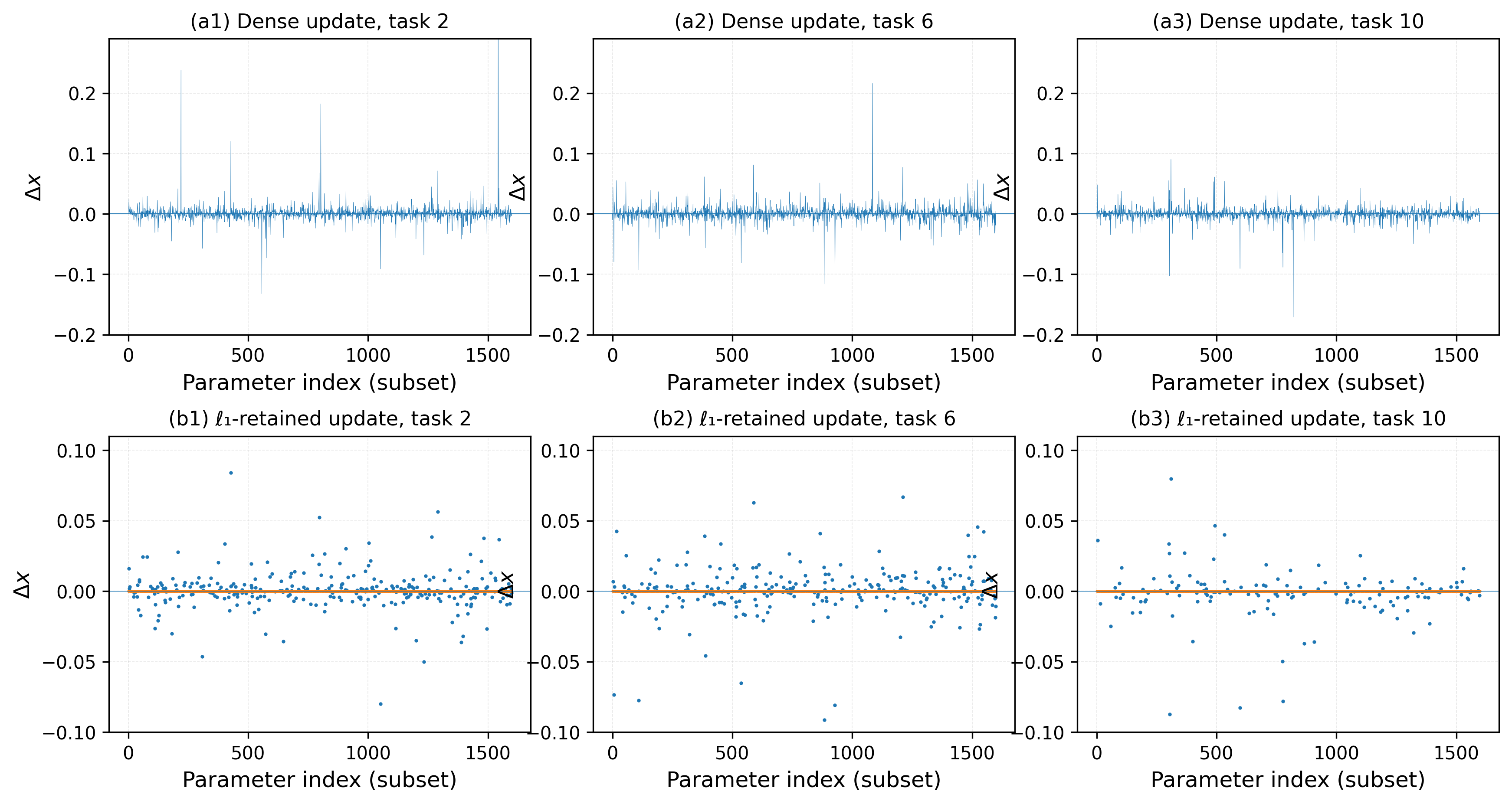}
    \caption{Evolution of parameter updates across tasks. The top row shows dense learning proposals from minimizing the current task loss, while the bottom row shows the corresponding updates after $\ell_1$-based selective retention. As the task index increases, updates become increasingly sparse, confining adaptation to a smaller subset of parameters.}
    \label{l1-loss}
\end{figure*} 
This stability is further illustrated in Fig. \ref{plas-stab}. In the left panel, we track the accuracy of the initial task ($T_1$) as subsequent tasks are introduced. While traditional regularization methods experience a significant decay in $T_1$ performance, our model maintains a stable profile. This validates our use of sparsity gate ($\tau$); that prevent the erasure of foundational knowledge. The right panel demonstrates that this stability does not come at the cost of plasticity, as our model achieves the highest accuracy across the 10-task sequence, outperforming state-of-the-art projection and meta-learning baselines.

\begin{figure}
    \centering
    \includegraphics[width=0.99\linewidth]{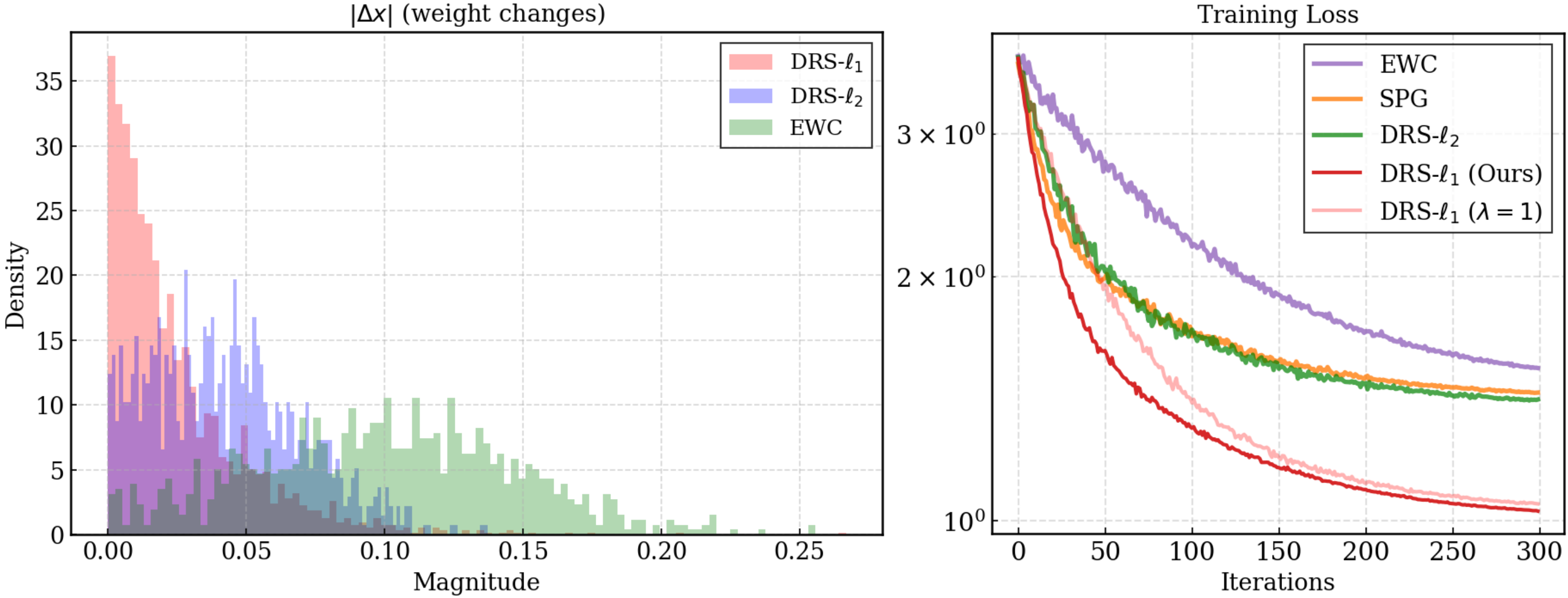}
    \caption{Comparison of parameter magnitude between DRS with \(\ell_1\) and \(\ell_2\), against EWC. Our proposed DRS-$l_1$ shows the best concentration near zero, indicating minimal disruption to pre-trained parameters. DRS-$l_2$ provides moderate regularization, while EWC allows the largest changes. }
    \label{dense}
\end{figure}

\vspace{-7pt}

\subsection{Selective Plasticity and Retention}
Fig.\ref{heat} visualizes the parameter update intensity across a 20-task sequence on CIFAR-100, comparing our model with EWC and SGD. Each heatmap illustrates the magnitude of parameter shifts, defined as $\Delta x=|x_t - x_{t-1}|$ per task, where bright regions denote large updates and dark blue is zero change. SGD is saturated with bright values, indicating that every new task overwrites the parameter space, which confirms the catastrophic forgetting seen in the stability plots (Fig. \ref{plas-stab}). While EWC reduces this intensity, it still shows low-magnitude updates across the network. Our heatmap is dominated by blue regions (zero update). By enforcing a threshold of $\tau$, the stability agent locks the majority of parameters, allowing only sparse, task-critical updates.

\subsection{Efficacy of Subspace Pruning}
The core component of our model is the enforcement of a sparsity gate through the $\ell_1$ proximal operator. Fig. \ref{dense} analyzes our model's stability and training convergence on CIFAR-100 (20 tasks) using a ResNet-18 backbone. In DRS-$L_1$ (ours), the density of weight changes is mostly concentrated at the origin ($|\Delta x| \approx 0$). Because, by enforcing $|\Delta x| = \text{prox}_{\gamma \lambda \|\cdot\|_1}(x)$, we create a dead zone where parameters with gradient updates below the threshold $\tau$ remain unchanged. DRS-$L_2$ and EWC demonstrate a wider spread. This is because $L_2$ applies a smooth penalty that lacks a thresholding mechanism. This widespread nudging of all parameters is the direct cause of the parameter drift that degrades accuracy across the task sequence. The training loss plot (right) highlights the efficiency of the decoupled DRS negotiation. Our model achieves the lowest training loss in fewer iterations compared to EWC, SPG, and even the $L_2$ variant of our model. This rapid descent proves that the reflection step ($2x-y$) can navigate the intersection between the new task loss and the stability constraint. Even when the regularization is reduced ($\lambda=1$), our model remains more efficient than the baselines, proving that the DRS itself is superior to standard coupled optimizers.

\subsection{Stability-Plasticity Analysis}
\begin{figure}
    \centering
    \includegraphics[width=1.01\linewidth]{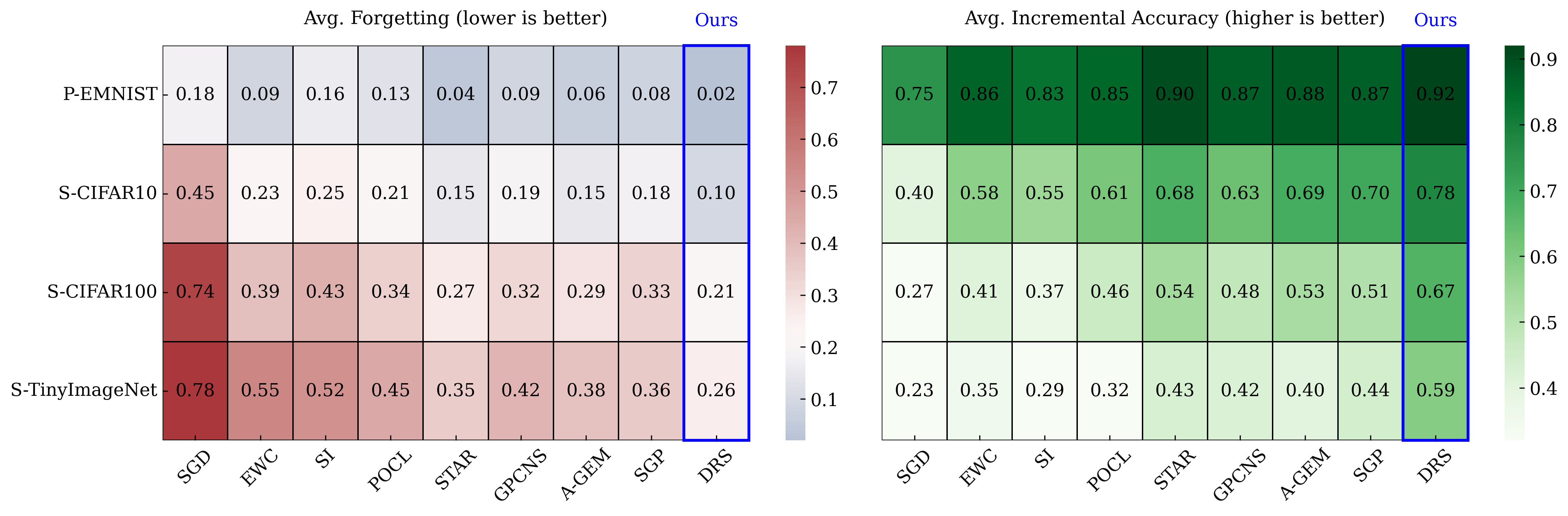}
    \caption{Stability-Plasticity in class-incremental learning. Most baselines (e.g., EWC, SI) fail to scale to complex manifolds (TinyImageNet), demonstrating both high forgetting and reduced plasticity. In contrast, our model decouples these objectives. It achieves retention (low forgetting) without compromising the learning potential (high plasticity).}
    \label{matrix}
\end{figure}

We further analyze the class-incremental learning using a ResNet-18 backbone to assess robustness against inter-task interference. While Table \ref{alexnet} reports the final snapshot accuracy, Fig. \ref{matrix} visualizes the maintenance of knowledge throughout the entire learning process. We report two metrics derived from the test accuracy matrix $R_{i,j}$ (accuracy on task $j$ after learning task $i$): Average forgetting, $F_T \downarrow$, measures the average performance drop on previously learned tasks by the end of training, $F_T=\frac{1}{T-1}\sum_{j=1}^{T-1}\max_{l<T}(R_{l,j}-R_{T,j})$; Average incremental accuracy ($A_{\text{Inc}} \uparrow$) that captures the area under the learning curve by averaging the mean accuracies at each incremental step $t$, $A_{\text{Inc}} = \frac{1}{T} \sum_{t=1}^{T} \left( \frac{1}{t} \sum_{j=1}^{t} R_{t,j} \right)$. High $A_{\text{Inc}}$ not only learn new tasks effectively (plasticity) but also maintains high performance on all preceding tasks at every intermediate step.  Baseline SGD displays severe forgetting on the S-TinyImageNet ($F_T=0.78$), resulting in a negligible incremental accuracy of $A_{\text{Inc}}=0.23$. This indicates that the learning of new task completely overwrites the parameter of previous tasks. While EWC reduces forgetting to $F_T=0.55$, its reliance on a rigid penalty limits the model's plasticity ($A_{\text{Inc}}=0.35$). Our model demonstrates superior stability-plasticity balance, achieving significantly lower forgetting ($F_T=0.26$) and the highest recorded incremental accuracy ($0.59$). The results confirm that the DRS framework outperforms even replay-based baselines like A-GEM and STAR without requiring the retention of raw training samples.

\vspace{-5pt}

\subsection{Long-Sequence Stability Analysis}
To evaluate the stability of our model against catastrophic forgetting, we conducted a large-scale class-incremental learning experiment using the CASIA-HWDB1.0 dataset \cite{liu2011casia} with ResNet-18 backbone. This benchmark is particularly challenging due to its fine-grained inter-class similarity. To construct a long-horizon learning scenario, we curated a subset of 3,700 classes (randomly discarding 55 rare classes from the original 3,755) and divided them into 100 disjoint tasks ($T=100$), with 37 classes per task.  The model is trained sequentially on $D_1, \dots, D_{100}$. To achieve a granular forgetting, we measure the group forgetting $F_{G}(t)$ for a specific task cluster $G$ (e.g., tasks 1-10) at step $t$: $F_{G}(t) = \frac{1}{|G|} \sum_{k \in G} \max_{j \in \{k, \dots, t-1\}} (A_{k,j} - A_{k,t})$, where $A_{k,j}$ is the accuracy on task $k$ after learning task $j$ (where $j \geq k$). This metric tracks the degradation of specific task as the model encounters new task distributions. Fig. \ref{forget} visualizes this through consecutive 10-task blocks and the cumulative average forgetting for all 100 tasks. As observed, while baseline models have a linear accumulation of forgetting, our DRS learner maintains a flat trend. 

%All models use a ResNet-18 backbone, we use a learning rate of $0.1$, decaying by a factor $0.1$ upon plateau. Batch size is set to 64. 

\vspace{-5pt}

\subsection{Plasticity and Network Stiffness Analyses}
\begin{figure}
    \centering
    \includegraphics[width=0.95\linewidth]{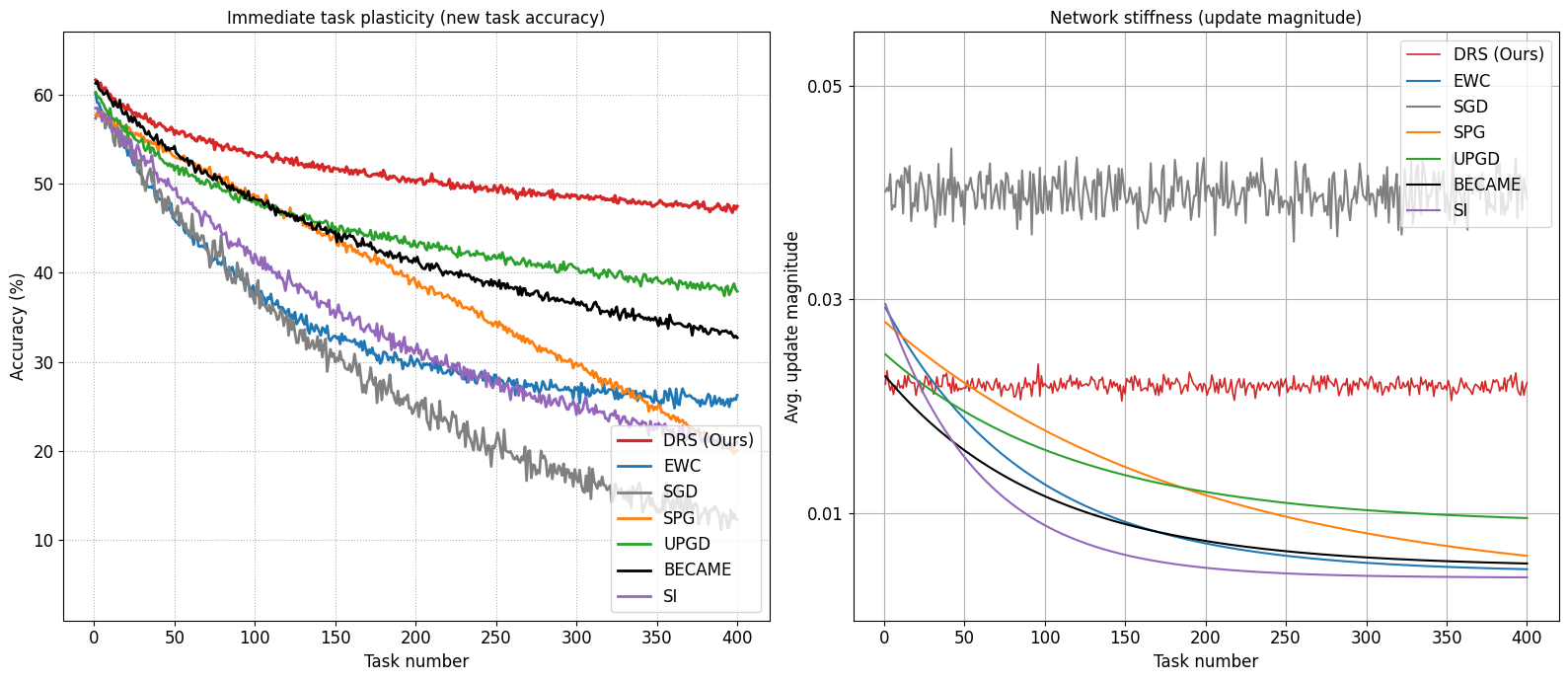}
    \caption{Long-horizon plasticity analysis. Our model maintains a consistent upward trend with reduced variance, showing its ability to preserve prior knowledge while effectively adapting to new tasks.  }
    \label{perm}
\end{figure}

To evaluate long-term learning, we conduct a 400-task experiment using a permuted label protocol (\cite{abbas2023loss, lyle2024disentangling, elsayed2024addressing, dohare2024loss}) on TinyImageNet. Unlike standard continual learning, which focuses on memory, this experiment assesses whether a model is able to adapt to new data after hundreds of previous tasks (which also known as maintaining plasticity). We simulate a continuous environment by cycling through Tiny-ImageNet,  where label mappings are randomly permuted every $\Delta=2,500$ steps. This creates extreme gradient interference, forcing the network to map identical visual features to different semantic categories sequentially. We then track task accuracy (learning capability) and network stiffness (average update magnitude $\Delta ||x||_2$). The results in Fig. \ref{perm} show most continual learning models fail in long sequences. In SGD and SI, plasticity drops sharply within the first 50 tasks. This is caused by weight saturation, where the cumulative gradients push parameters into dead zones of the loss landscape, making them unresponsive to new task signals. Moreover, as shown in Fig. \ref{perm}(right), EWC, SI and BECAME suffer from a stiffening effect where the update magnitude approaches zero. This indicates a total loss of plasticity; the constraints have effectively frozen the parameters, preventing the model from adapting to more number of tasks. SGD stays high and erratic. Because it lacks stabilization, moving every weight and overwriting old knowledge. Our model maintains a stable, non-zero profile. While its magnitude is lower than SGD, reflecting the efficiency of our $\ell_1$ sparsity gate. Indeed, the reflection step ($2x-y$) in the DRS solver acts as a buffer for gradient conflict. Instead of rigidly locking every weight, the decoupled negotiation allows the model to reconfigure the feature manifold without destroying the past representations, ensuring the network is functional across long-task horizon.

\begin{figure*}
    \centering
    \includegraphics[width=0.89\linewidth]{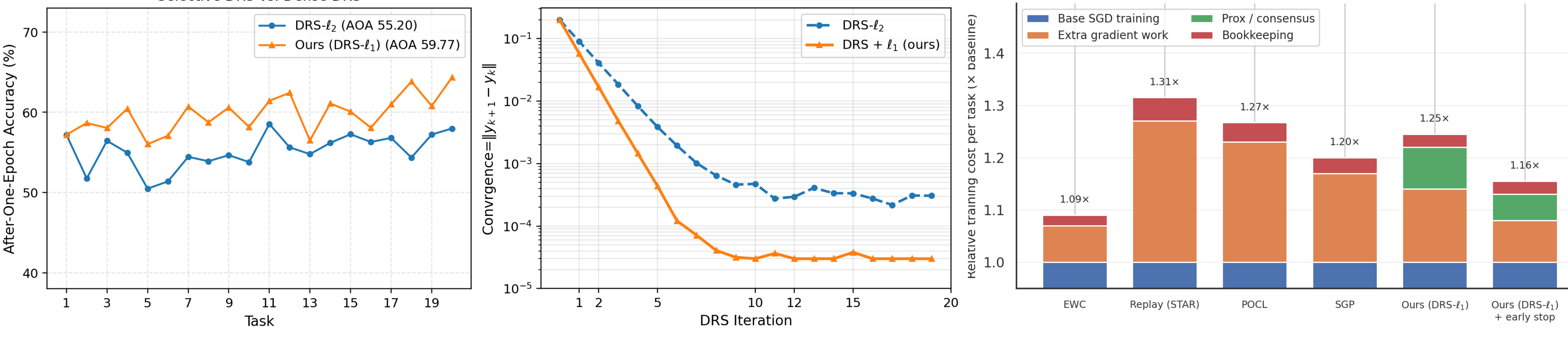}
    \caption{Selective DRS vs. Dense DRS on the fine-grained CUB-200 dataset. (Left) We report after-one-epoch accuracy across tasks. Both models start from the same accuracy at task 1, as no forgetting has yet occurred. As more tasks are added, DRS-$\ell_1$ maintains a higher AOA than DRS-$\ell_2$. The middle panel shows the convergence residual across DRS iterations. Our model converges faster and to a lower point, and efficiently solve conflict between task loss and stability constraints. The right panel decomposes per-task training cost relative to standard SGD. Our model introduces only a marginal computational overhead ($1.16\times$) due to inexpensive proximal and bookkeeping operations. }
    \label{cost}
\end{figure*}

\begin{figure}
    \centering
    \includegraphics[width=0.95\linewidth]{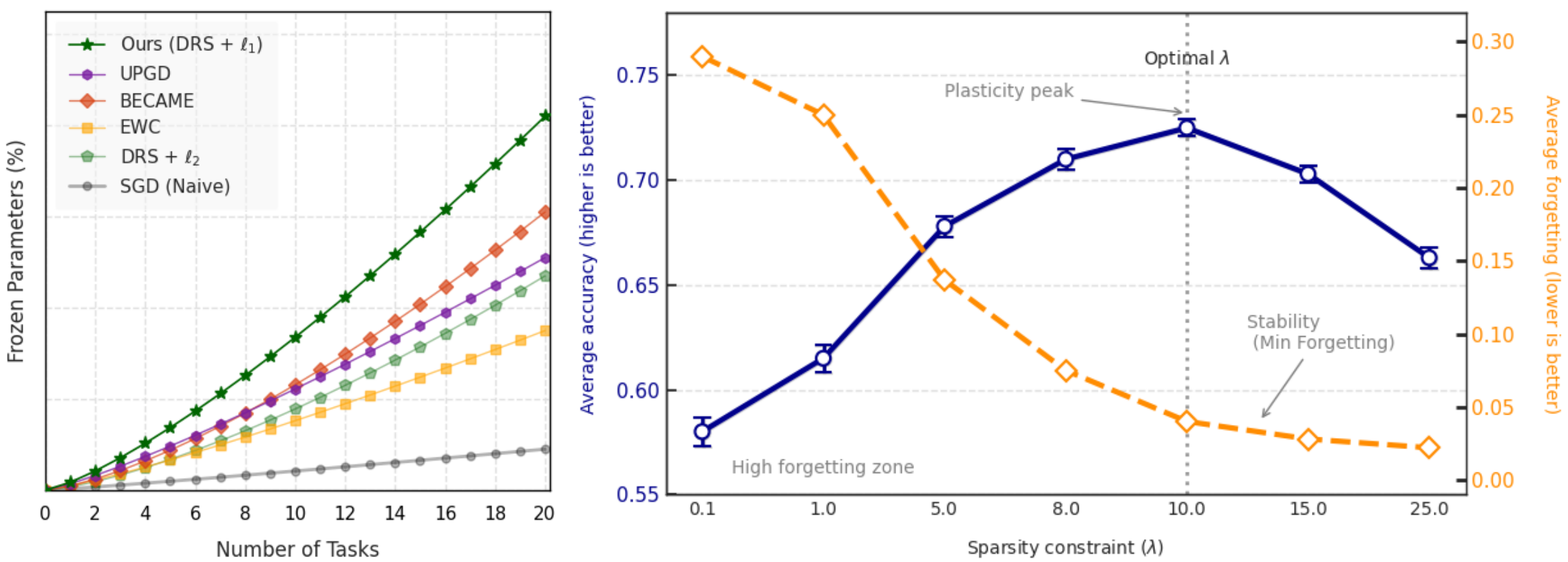}
    \caption{Effect of sparsity constraint $\lambda$ on average accuracy and forgetting on CIFAR100 data. Forgetting decreases with larger, but with diminishing adaptation after $\lambda > 10$.}
    \label{abla-1}
\end{figure}
% %%%%%%%%%%%%%%%%%%%%%%
% \begin{figure}
%     \centering
%        \includegraphics[width=0.85\linewidth]{iclr2026/CF.pdf} \vspace{-10pt}
%     \caption{................... } 
%         \label{cf}
% \end{figure}
% %%%%%%%%%%%%%%%%%%%%%%

\vspace{-5pt}

%%%%%%%%%%%%%%%%%%%%%%%%%%%%%%
\subsection{Ablation Studies}

\noindent {\bf{Effect of structural sparsity: } }
The effective stability is evaluated across 20 sequential tasks on split CIFAR-100 in Fig. \ref{abla-1}, using ResNet-18. We define effective stability as the percentage of parameters that remain stationary ($\frac{|\Delta w|}{|w|} < \epsilon$) after learning a new task. By identifying and freezing $\approx 83\%$ of the network by task 20, our model (DRS + $\ell_1$) protects the representations of prior tasks. Unlike EWC, which treats the regularization as a smooth gradient, the DRS solver uses the proximal operator and solves the inclusion condition $0 \in \nabla \mathcal{L} + \partial \|w\|_1$.  This creates a hard filter; it evaluates the learning proposal generated by the task loss and forces non-essential parameters to zero (i.e., $\epsilon\approx0.05$). This prevents the gradient drift observed in the SGD and EWC, where unused parameters fluctuate randomly due to stochastic noise ($\nabla \mathcal{L} \neq 0$), and leads to catastrophic forgetting. While UPGD also employs hard constraints via gradient projection, it is lower than our model by a margin of $\approx 12-34\%$. Our model locks low-level features in early tasks (tasks 1-8), requiring fewer parameter updates as the sequence progresses. %%For EWC, as the number of tasks increases ($T \to 20$), the magnitude of the new task's loss often overwhelms the accumulated regularization terms. This results in the high-plasticity gradients from the new task overriding the soft penalty constraints, causing previously stable parameters to break loose and drift. 

\noindent {\bf{Interpretation of selective plasticity:} }
Our model decouples learning and retention by separating the dense learning proposal from the sparse retention step. At each task $t$, minimizing the task loss produces a dense learning proposal in parameter space. This proposal is defined by the shift between the proximal update and the consensus point, as $\Delta x\!=\!y_{k+1}-y_k$. The $\ell_1$ proximal operator acts as a sparsity gate. It evaluates the proposal and forces non-essential parameter updates to zero while allowing critical updates to proceed. Fig. \ref{l1-loss} visualizes the evolution of these parameter updates across a sequence of tasks on CIFAR-100 using ResNet-18.  The top row confirms that plasticity is not suppressed throughout the task sequence, while the bottom row shows the effective updates after applying the proposed $\ell_1$-based retention mechanism. As more tasks are learned, parameters that are important for previous tasks receive a larger weight $F_i$. The $\ell_1$ regularizer penalizes these important weights, suppressing any new updates that would cause forgetting. Parameters with low importance weights remain free to adapt to the current task. The figures confirm that learning continues with a consistent update magnitude but is restricted to a smaller subset of parameters.

\noindent {\bf{Convergence analysis of the DRS solver:} }  
The DRS solver seeks a fixed point where the auxiliary sequence $y_k$ stabilizes, meaning $y_{k+1} \to y_k$. We monitor this progress using the residual $r_k = \|y_{k+1} - y_k\|$, which measures the agreement between plasticity (task loss) and stability (regularization) updates. As shown in Fig. \ref{cost}(Left), both $\ell_1$ and $\ell_2$ confirm the stable convergence, but the $\ell_1$-based solver converges faster and reaches a lower residual within fewer iterations. The  $\ell_2$ variant converges more gradually because it preserves dense parameter updates, which slows fixed-point stabilization. We evaluate the real-world impact of this convergence on the CUB-200 dataset across 20 sequential tasks. We compare dense DRS ($\ell_2$) against our selective DRS ($\ell_1$) in terms of After-One-Epoch Accuracy (AOA). Both models achieve identical accuracy on the first task, as no stability exist yet. As the task sequence progresses, dense DRS provides performance fluctuations and lower AOA, indicating that uniform $\ell_2$ shrinkage restricts adaptation even for task-irrelevant parameters. In contrast, our model maintains higher AOA throughout the task sequence. 

%\vspace{-5pt}

\subsubsection{Computational Cost}
We analyze the computational overhead of the proposed learner on CUB-200 (20 tasks) using a ResNet-18 backbone, as shown in Fig.\ref{cost}(Right). While our model uses inner splitting iterations to balance plasticity and stability, the total training cost is slightly longer to simpler baselines  (EWC, SI). Unlike Bayesian methods that require expensive variational inference or Meta-learning approaches that rely on second-order derivatives (Hessians), each DRS iteration is computationally thin. The proximal operator for the $\ell_1$ is a soft-thresholding, which admits a closed-form solution, and its computation is a simple element-wise comparison with $O(d)$ complexity, where $d$ is the number of parameters.  Indeed, a unique advantage of the DRS solver is its rapid stabilization of the auxiliary sequence $y_k$. By monitoring the residual $\|y_{k+1}-y_k\|$, we apply an early-stopping rule that reduces the average iteration count from 9 to 6 per task. This optimization lowers the training time to $\sim1.16\times$ of the SGD baseline, but with significantly higher accuracy retention. Importantly, our model has zero inference overhead. Because DRS operates within the optimization phase to find a sparse parameter configuration, it does not alter the final network architecture or introduce task-specific modules.

\vspace{-5pt}

\section{Conclusion}
Continual learning is often described as a tension between stability and plasticity, where improving one dimension comes at the expense of the other. Most existing solutions address this by modifying the model itself, such as, storing past examples, expanding architectures, or introducing complex training schemes.  In this work, we took a different route and revisit the optimization process instead. By formulating task learning and knowledge retention as two complementary operators and solving them jointly using a Douglas-Rachford scheme, we showed that sparse, parameter-level selectivity can be enforced without replay buffers or architectural changes. The resulting learner achieves strong continual-learning performance across long task sequences while remaining within a simple regularization-based framework.

%%%%%%%%%%%%%%%%%%%%%%%%%%%%%%%%%%

%%%%%%%%%%%%%%%%%%%%%%%%%%%%%%%%%%%%%

\bibliographystyle{unsrt}
\bibliography{template}

%%%%%%%%%%%%%%%%%%%%%%
%\vspace{-2cm}
\begin{IEEEbiography}
[{\includegraphics[width=0.85in,height=0.85in,clip,keepaspectratio]{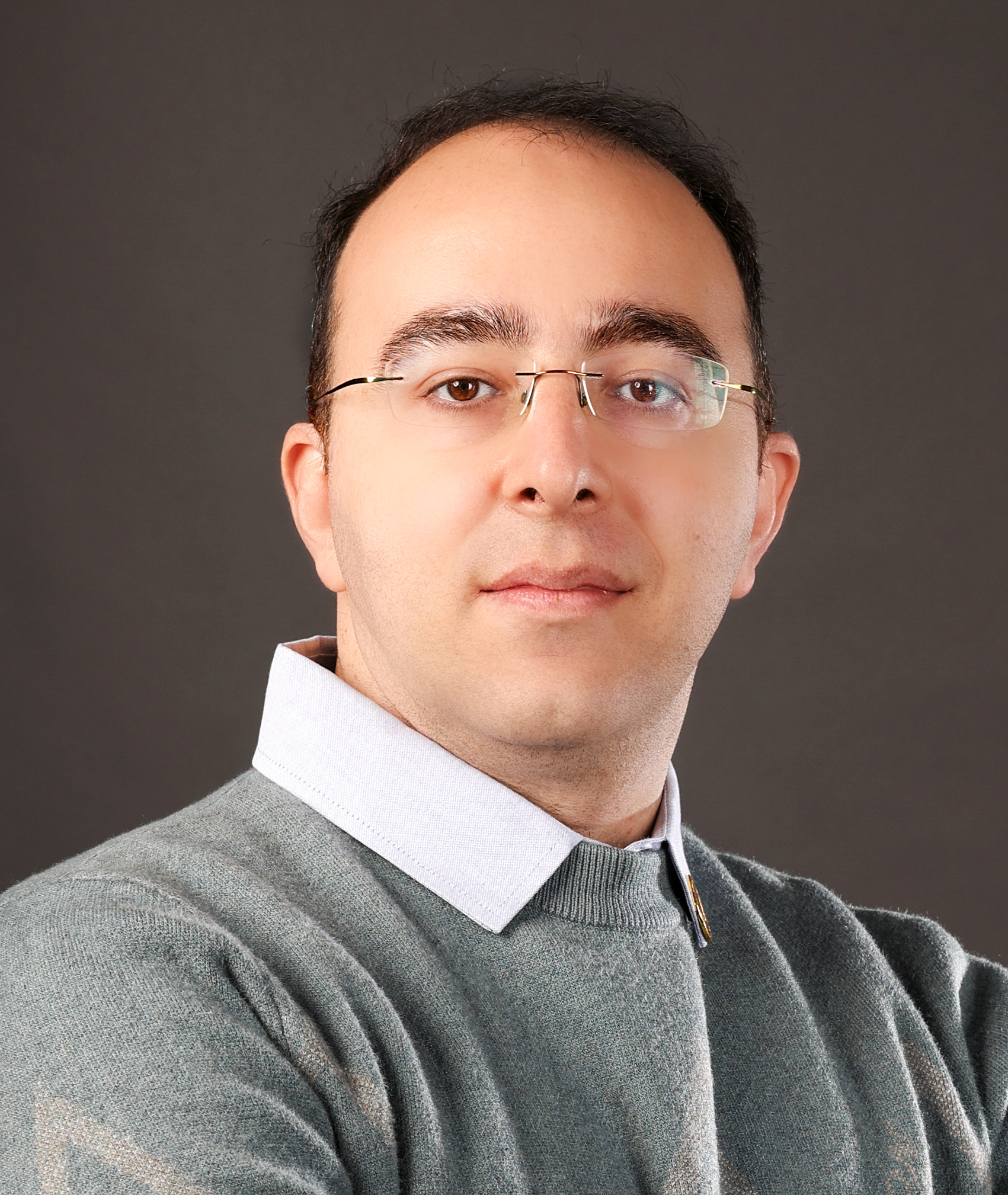}}]
{Pourya Shamsolmoali} (Senior Member, IEEE) received the PhD degree in computer science from Shanghai Jiao Tong University. He has been a visiting researcher with Linköping University, INRIA-France, and ETS-Montreal. He is currently a lecturer at the University of York. His main research focuses on machine learning, computer vision, and image processing.
\end{IEEEbiography}
\vspace{-2cm}
%%%%%%%%%%%%%%%%%%%%%%%
\footnotesize{
\begin{IEEEbiography}
[{\includegraphics[width=0.85in,height=0.85in,clip,keepaspectratio]{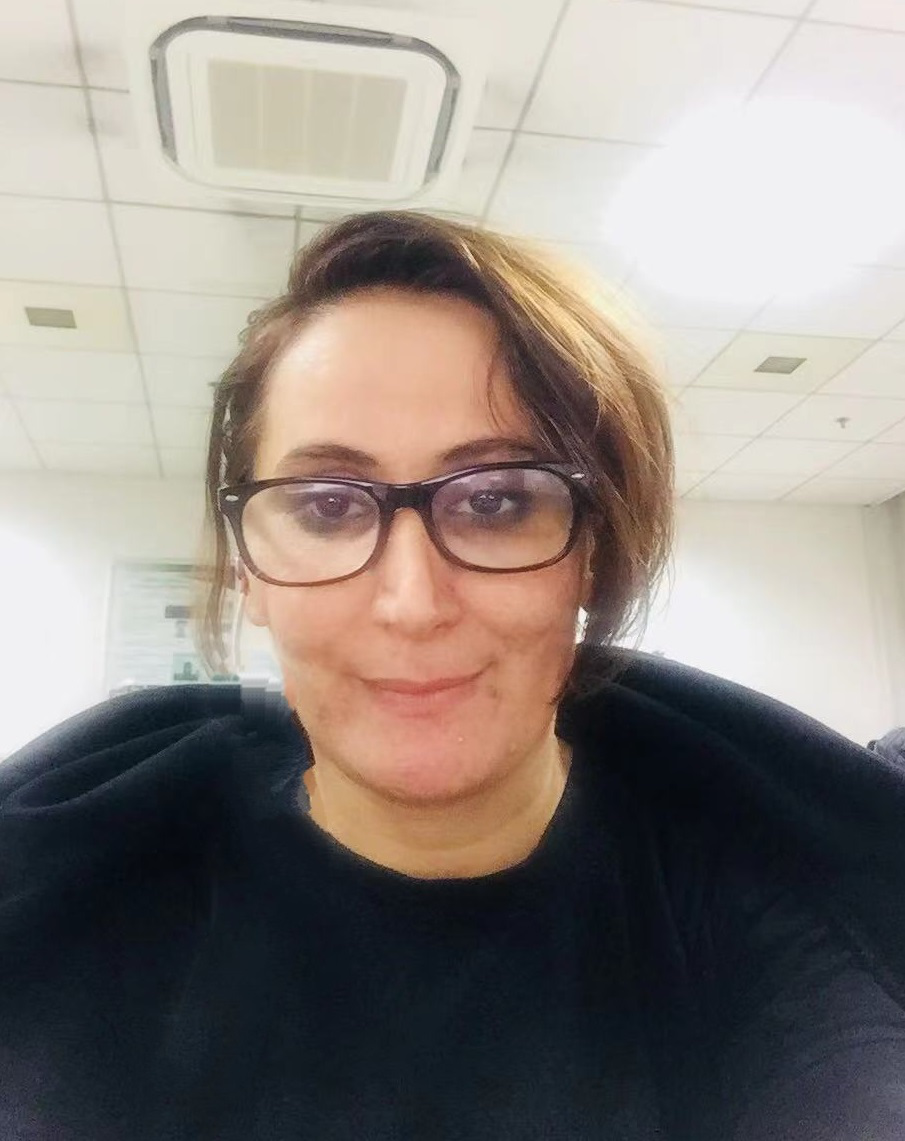}}]
{Masoumeh Zareapoor} (Member, IEEE) received the PhD degree in computer science from Jamia University, India. She is an associate researcher at Shanghai Jiao Tong University and was a research engineer at Huawei and NWPU Xi'an. Her main research focuses on machine learning and computer vision.
\end{IEEEbiography}}
\vspace{-2cm}
%%%%%%%%%%%%%%%%%%%%%%%%%%
%%%%%%%%%%%%%%%%%%%%%%%
\footnotesize{
\begin{IEEEbiography}
[{\includegraphics[width=0.85in,height=0.85in,clip,keepaspectratio]{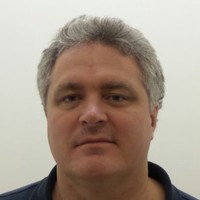}}]
{Eric Granger} (Member, IEEE) received a Ph.D. degree in electrical engineering from Ecole Polytechnique de Montreal, Canada, in 2001. 
%He was a Defense Scientist with DRDC Ottawa from 1999 to 2001, and in Research and Development with Mitel Networks from 2001 to 2004. 
He joined the Dept. of Systems Engineering at École de technologie supérieure (ÉTS) in Montréal, in 2004, where he is currently a Professor and the Director of LIVIA. He is ETS Industrial Research Co-Chair in Embedded Neural Networks for Intelligent Connected Buildings (Distech Controls Inc.) and former FRQS Co-Chair in AI and Health (2021-24), with research interests in pattern recognition, machine learning, information fusion, and computer vision.
\end{IEEEbiography} 
%%%%%%%%%%%%%%%%%%%%%%%%%%
\vspace{-2cm}
%%%%%%%%%%%%%%%%%%%%%%%
\footnotesize{
\begin{IEEEbiography}
[{\includegraphics[width=0.85in,height=0.85in,clip,keepaspectratio]{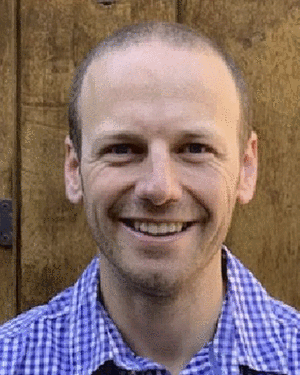}}]
{William A. P. Smith} received the BSc degree in computer science and the PhD degree from the University of York. He is currently a Professor with the Department of Computer Science, University of York. He holds a Royal Academy of Engineering/The Leverhulme Trust Senior Research Fellowship. His research focuses on shape and appearance modelling and model-based supervision.
\end{IEEEbiography}}
\vspace{-2cm}
%%%%%%%%%%%%%%%%%%%%%%%%%%
\footnotesize{
\begin{IEEEbiography}
[{\includegraphics[width=0.85in,height=0.85in,clip,keepaspectratio]{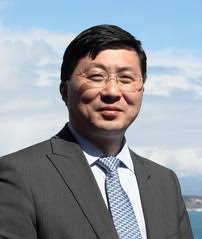}}]
{Yue Lu} (Senior Member, IEEE) is a Professor and Dean of the School of Communication and Electronic Engineering at East China Normal University. He received his B.S. and M.S. from Zhejiang University in telecommunications and electronic systems, respectively, and his Ph.D. in pattern recognition and intelligent systems from Shanghai Jiao Tong University. Previously, he was a research fellow at the National University of Singapore, department of computer science.
%He has over 150 publications, 19 authorized patents, and is an Associate Editor of Pattern Recognition.
\end{IEEEbiography}}
\vspace{-2cm}
%%%%%%%%%%%%%%%%%%%%%%%%%%

\end{document}